\documentclass[10pt,twocolumn,letterpaper]{article}
\usepackage[pagenumbers]{cvpr} 

\usepackage{graphicx}
\usepackage{amsmath}
\usepackage{amssymb}
\usepackage{booktabs}
\usepackage[toc,page]{appendix}

\usepackage{multirow}
\usepackage{pifont}
\usepackage{amssymb}
\usepackage{subfloat}
\usepackage{algorithm}
\usepackage{algorithmic}
\usepackage{amsfonts}
\usepackage{amsmath}
\usepackage{wrapfig}
\usepackage{graphicx}
\usepackage[vlined,algo2e]{algorithm2e}
\SetCommentSty{mycommfont}
\SetKwInput{KwInput}{Input}                
\SetKwInput{KwOutput}{Output}              
\SetKwProg{Fn}{Function}{:}{}
\usepackage{booktabs}
\newcommand{\tabincell}[2]{\begin{tabular}{@{}#1@{}}#2\end{tabular}}
\usepackage{multirow}
\usepackage{pifont}

\usepackage{setspace}

\newcommand{\cmark}{\ding{51}}%
\newcommand{\xmark}{\ding{55}}%
\newcommand*{\affaddr}[1]{#1} 

\newcommand*{\email}[1]{\texttt{#1}}
\usepackage[pagebackref,breaklinks,colorlinks]{hyperref}

\usepackage[capitalize]{cleveref}
\crefname{section}{Sec.}{Secs.}
\Crefname{section}{Section}{Sections}
\Crefname{table}{Table}{Tables}
\crefname{table}{Tab.}{Tabs.}

\begin{document}

\title{Neural Collaborative Graph Machines for Table Structure Recognition}

\author{%
	Hao Liu\textsuperscript{\dag\thanks{Equal contribution. \textsuperscript{\dag}Contact person.}} \quad Xin Li\textsuperscript{*} \quad Bing Liu \quad Deqiang Jiang \quad Yinsong Liu \quad Bo Ren\\		
	\affaddr{Tencent YouTu Lab} \quad\\
	\email{\small \{ivanhliu, fujikoli, billbliu, dqiangjiang, jasonysliu, timren\}@tencent.com}
}
\maketitle

\begin{abstract}
	Recently, table structure recognition has achieved impressive progress with the help of deep graph models. Most of them exploit single visual cues of tabular elements or simply combine visual cues with other modalities via early fusion to reason their graph relationships. However, neither early fusion nor individually reasoning in terms of multiple modalities can be appropriate for all varieties of table structures with great diversity. Instead, different modalities are expected to collaborate with each other in different patterns for different table cases. In the community, the importance of \textit{intra-inter} modality interactions for table structure reasoning is still unexplored. In this paper, we define it as \textit{heterogeneous} table structure recognition~(Hetero-TSR) problem. With the aim of filling this gap, we present a novel Neural Collaborative Graph Machines (NCGM) equipped with stacked collaborative blocks, which alternatively extracts intra-modality context and models inter-modality interactions in a hierarchical way. It can represent the intra-inter modality relationships of tabular elements more robustly, which significantly improves the recognition performance.  We also show that the proposed NCGM can modulate collaborative pattern of different modalities conditioned on the context of intra-modality cues, which is vital for diversified table cases. Experimental results on benchmarks demonstrate our proposed NCGM achieves state-of-the-art performance and beats other contemporary methods by a large margin especially under challenging scenarios.   
\end{abstract}

\begin{figure}[htb]
	\centering
	\includegraphics[width=1\linewidth, height=5.5cm]{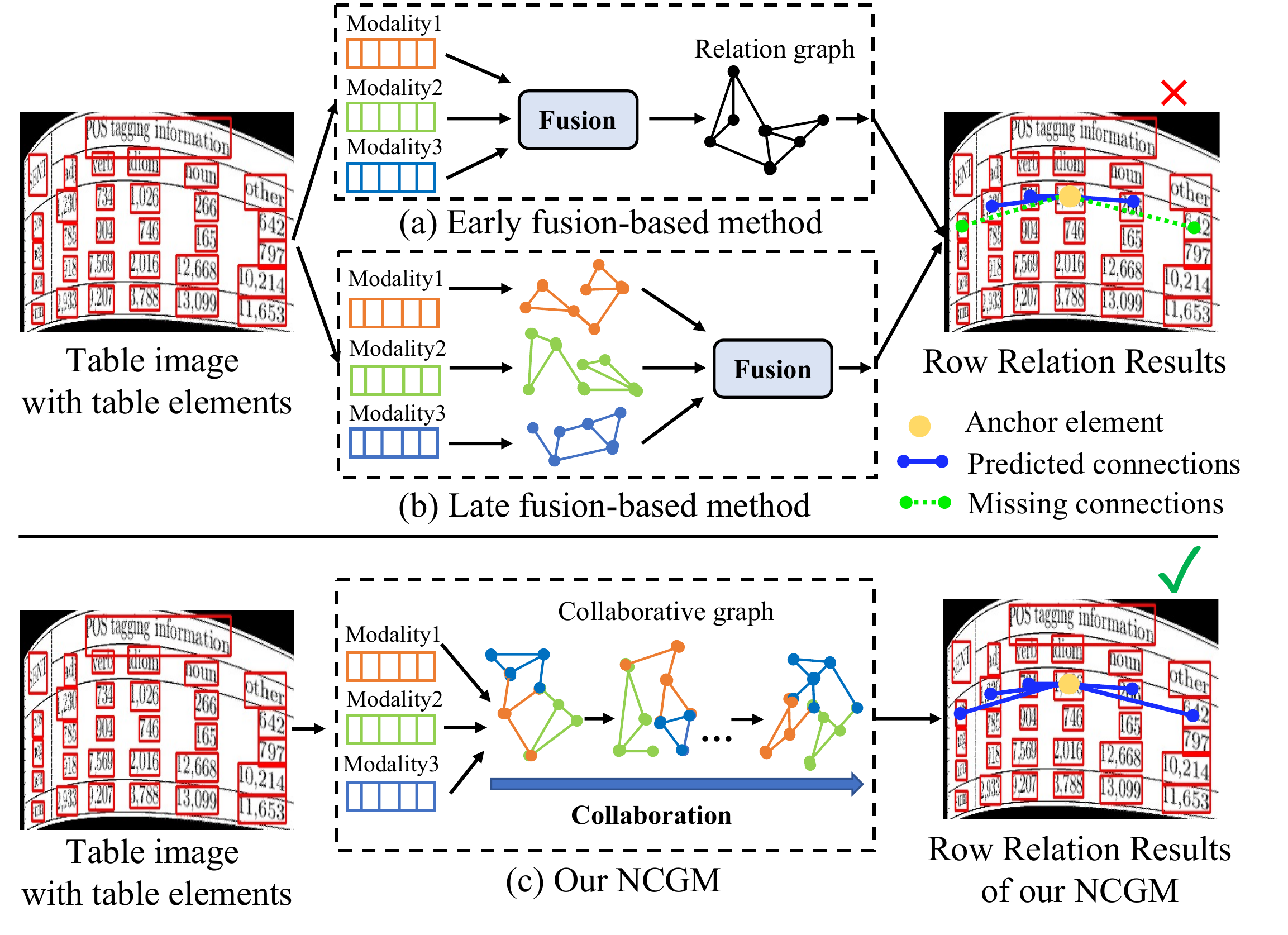}
	\vspace{-7mm}
	\caption{{Illustration of motivation of the proposed NCGM. (a)~Early fusion-based method. The multiple modalities of table elements are fused before modeling their relationships. (b)~Late fusion-based method. The multiple modalities are modeled on their intra-modality relationships which are then fused for final results prediction. Due to lack of collaboration, for a distorted table case, previous methods cannot well extract the row relations~(connected by blue lines) for an anchor element~(yellow) with some true relation lost~(green dotted line). (c)~Our proposed NCGM. Different modalities are built into graphs with collaboration, which well accommodate the distorted table case.}} 
	\label{fig:moti}
	\vspace{-5mm}
\end{figure}
\section{Introduction}
	Table structure recognition (TSR) aims to recognize the table internal structure to the machine readable data mainly presented in two formats: \textit{logical structure}~\cite{li2020tablebank,end2end1} and \textit{physical structure}{~\cite{schreiber2017deepdesrt,paliwal2019tablenet,khan2019table,tensmeyer2019deep,chi2019complicated,rethinkingGraphs,raja2020table,zheng2021global,qiao2021lgpma,liu2021show,long2021parsing}}. More concretely, logical structure only focuses on whether two table elements belong to the same row, column or cells (\textit{i.e.}, logical relationships), while the physical one contains not only logical relationships but also physical coordinates of cell boxes. The recognized tabular structure is essential to many downstream applications~\cite{jauhar2016tables, li2016deep}. Although many previous  algorithms{~\cite{li2020tablebank,end2end1,schreiber2017deepdesrt,khan2019table,tensmeyer2019deep,chi2019complicated,rethinkingGraphs,raja2020table,zheng2021global,qiao2021lgpma,liu2021show,long2021parsing}} have achieved impressive progress in the community, TSR is still a challenging task due to two factors of complicated tables. The interior factor is complex table structure where spanning cell occupies at least two columns or rows, while exterior one is table distortion incurred by capture device.
	
	Intuitively, \textit{table elements} (text segment bounding boxes or table cells) commonly have inherent relationships and natural graph structure. Therefore, recent methods~\cite{chi2019complicated,rethinkingGraphs,raja2020table} attempt to attack the problem via constructing visual cues of table elements as graphs and applying the deep graph model, such as Graph Convolutional Networks (GCN)~\cite{kipf2016semi} to reason their relationships. To introduce richer table information, several methods~{\cite{rethinkingGraphs,raja2020table,liu2021show}} concatenate the visual features with other modalities of features, such as geometry features, as a whole input to the graph model, as shown in Fig.~\ref{fig:moti}~(a). Nevertheless, the relational inductive biases of different modalities would be highly discrepant, which makes naively early-fused modalities unable to deal with all table structures of great diversity. Besides, the intra-modality relationships would negatively affect each other when reasoning specific table structures. For example, the coordinates of table would dominate when recognizing a regular table, but they would become unreliable when processing distorted table cases. Instead, another alternative way is to individually model intra-modality relationships between table elements and combine them by a late-fusion strategy~(Fig.~\ref{fig:moti}~(b)). Unfortunately, the disentangled reasoning in terms of intra-modality interactions would introduce the curtailment of inter-modality interactions. This dilemma leads to the following question: \textit{can different modalities collaborate with each other rather than interfering under different table scenarios?} We define this practical problem as heterogeneous table structure recognition~(Hetero-TSR), which still lacks investigation.
	
	In this work, we propose a novel Neural Collaborative Graph Machines (NCGM) tailored for this problem, as illustrated in Fig.~\ref{fig:moti}~(c). Concretely, we adopt text segment bounding boxes as table elements in our method and extract their multi-modality feature embeddings from appearance, geometry and content dimensionality separately. To obtain the corresponding graph context and explore their interactions, we go beyond the standard attention model and propose a basic collaborative block with two successive modules,\textit{ i.e., }Ego Context Extractor~(ECE) and Cross Context Synthesizer~(CCS). Among, ECE plays a role that dynamically generates graph context for the samples of each modality while the subsequent CCS is in charge of fusing and modulating inter-modality interactive information for different table cases. We stack this elemental block multiple times. Through this way, the intra-modality context generation and inter-modality collaboration can be conducted alternatively in a hierarchical way, which enables intra-inter modality interactions to be generated constantly from the low layer to the top one. In other words, the low-level contextual information in multiple modalities and the high-level one can collaborate with each other throughout the whole network, which is similar to the human perception process~\cite{anderson2005cognitive,olshausen1993neurobiological}. The yielded collaborative graph embeddings enable our method to achieve better performance compared to other TSR methods, especially under more challenging scenarios, as clearly validated by extensive experimental results. To sum up, our contributions are in the four folds: 

\begin{itemize}
	\item We investigate the importance of collaboration between different modalities in TSR and propose the Hetero-TSR problem. To our best knowledge, we are the first to research the collaborative patterns between modality interaction for predicting table structure.
	\vspace{-2mm}
	\item We coin a novel NCGM tailored for Hetero-TSR problem, which consists of collaborative blocks alternatively conducting intra-modality context extraction and inter-modality collaboration in a hierarchical way. 
	\vspace{-2mm}
	\item Experimental results on public benchmarks demonstrate that our method significantly outperforms the state-of-the-arts.
	\vspace{-2mm}
	\item We release a synthesizing method to augment existing benchmarks to more challenging ones. Under more challenging scenarios, our method can achieve at most {11\%} improvement than the second best method.
\end{itemize}

\begin{figure*}[htb!]
	\centering
	\includegraphics[width=0.96\linewidth, height=3.6cm]{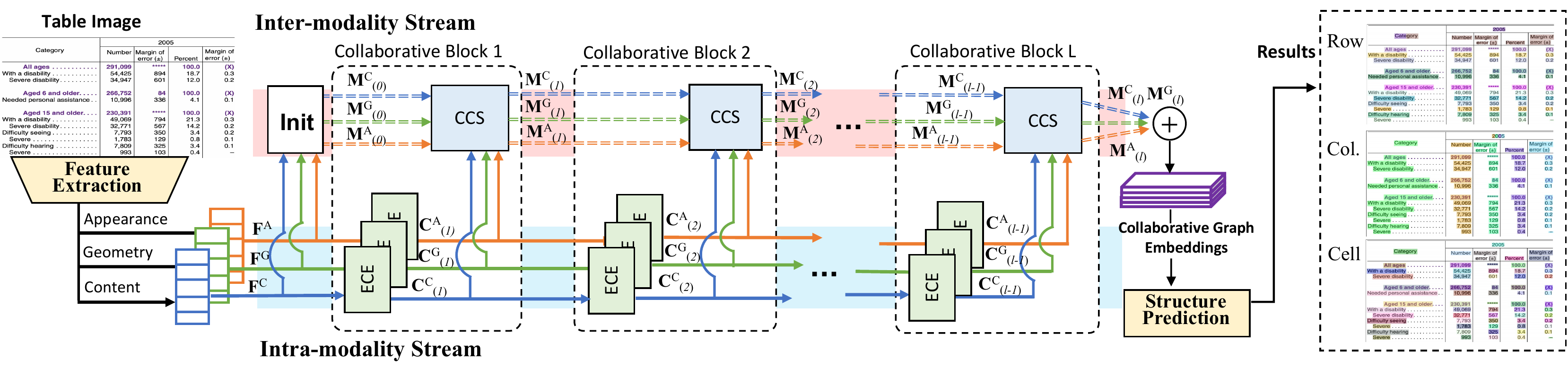}
	\vspace{-4mm}
	\caption{The architecture of our proposed method. Best viewed in color.} 
	\label{fig:arch}
	\vspace{-5mm}
\end{figure*}
\section{Related Work}
\subsection{Table Structure Recognition}
Before the flourishing of deep learning, traditional table structure recognition methods rely on pre-defined rules and hand-crafted features~\cite{itonori1993table,green1995recognition, hirayama1995method, tupaj1996extracting, kieninger1998t}. With the development of deep learning, table structure recognition methods have recently advanced substantially on performance, which can be classified into three categories: boundary extraction-based~\cite{schreiber2017deepdesrt, paliwal2019tablenet, khan2019table, tensmeyer2019deep,long2021parsing}, generative model-based~\cite{li2020tablebank, end2end1}, and graph-based~\cite{chi2019complicated, rethinkingGraphs, raja2020table,liu2021show} methods.

\vspace{-2mm}
\paragraph{Boundary extraction-based methods.} To extract cell boundaries, DeepDeSRT~\cite{schreiber2017deepdesrt} and TableNet~\cite{paliwal2019tablenet} are proposed by utilizing semantic segmentation. 
Besides, another technique~\cite{khan2019table} exploits bi-directional GRUs to establish row and column boundaries in a context driven manner. However, these methods are struggled when identifying cells spanning multiple rows and columns. SPLERGE~\cite{tensmeyer2019deep} splits the table into grid elements in which adjacent ones are merged to restore spanning cells, whereas it still suffers from boundary ambiguity problem. To tackle this issue, the hierarchical GTE~\cite{zheng2021global} leverages
clustering algorithm for cell structure recognition. {Cycle-CenterNet~\cite{long2021parsing} exploits the cycle-pairing module to simultaneously detect and group tabular cells into structured tables, which focuses on the precision of cell boundary of the wired table in the wild.}
In the similar spirit, LGPMA~\cite{qiao2021lgpma} applies soft pyramid mask learning mechanism on both the local and global feature maps. Nevertheless, the subsequently heuristic structure recovery pipeline cannot achieve decent performance in complex scenarios.

\vspace{-2mm}
\paragraph{Generative model-based methods.}
The method~\cite{li2020tablebank} utilizes the encoder-decoder framework, which generates an HTML tag sequence that represents the arrangement of rows and columns as well as the type of table cells.
Moreover, another generative algorithm~\cite{end2end1}, termed EDD, consists of an encoder, a structure decoder and a cell decoder.
The encoder captures visual features of input table images, while the structure decoder reconstructs table structure and helps the cell decoder to recognize cell content.

\vspace{-2mm}
\paragraph{Graph-based methods.}
GraphTSR~\cite{chi2019complicated} employs graph attention blocks to learn the vertex and edge representations in the latent space, and classifies edges as horizontal, vertical or unrelated. 
The method~\cite{rethinkingGraphs} introduces DGCNN to predict the relationship between words represented by the appearance and geometry features.
Also based on DGCNN, TabStruct-Net~\cite{raja2020table} proposes an end-to-end network training cell detection and structure recognition networks in a joint manner. {Besides, FLAG-Net~\cite{liu2021show} leverages the modulatable dense and sparse context of table elements. However, the above graph-based works are mostly designed for the interaction between table elements but lack the cues of the collaborative pattern of different modalities. In contrast to these works, our proposed NCGM leverages modality interaction to boost the multimodal representation for complex scenarios.}

\subsection{Transformer-based Multimodal Fusion}
Transformer~\cite{vaswani2017attention} architecture not only achieves significant performance gains in NLP community~\cite{devlin2018bert,liu2019roberta,lan2019albert,raffel2019exploring,sun2019ernie}, but also gives birth to several pre-training methods~\cite{lu2019vilbert,xu2020layoutlm,li2021selfdoc} fusing various modalities for multimodal tasks.

\vspace{-4mm}
\paragraph{Multiple embeddings fusion.}
VL-BERT~\cite{su2019vl} inheriting from BERT~\cite{devlin2018bert} introduces additional visual feature embeddings for visual-linguistic representations.
LayoutLM~\cite{xu2020layoutlm} is a document understanding pre-trained model, which jointly models the interactions between text and layout information across scanned document images.
However, the above algorithms simply take early-fused multiple embeddings as inputs, which may ignore the interactions between different modalities and result in discretization error and important details missing.

\vspace{-4mm}
\paragraph{Co-attentional fusion.}
To better utilize visual-linguistic representations, ViLBERT~\cite{lu2019vilbert} processes both visual and textual inputs in separate streams that interact through co-attentional transformer layers. Moreover, SelfDoc~\cite{li2021selfdoc} establishes the contextualization over a block of content via cross-modal learning to manipulate visual features and textual features. Nevertheless, these previous co-attention based methods can only handle two modalities. By comparison, our proposed NCGM focuses on modality collaboration rather than simple fusion. Further, NCGM can not only process the interaction among more than two individual modalities, but also alternatively conduct intra-modality context extraction and inter-modality collaboration, which exploits more useful information provided by different modalities.

\vspace{2mm}
\section{Methodology}
\subsection{Overall Architecture}\label{overall}
The overview of the proposed Neural Collaborative Graph Machines (NCGM) is shown in Fig.~\ref{fig:arch}. It mainly consists of collaborative blocks, which have two successive Multi-head Attention-based~\cite{vaswani2017attention} modules,  \textit{i.e.}, Ego Context Extractor~(ECE) and the Cross Context Synthesizer~(CCS). First, three modalities of feature embeddings~($\mathbf{F}^\sim \in \left\lbrace\mathbf{F}^{\text{G}},  \mathbf{F}^{\text{A}}, \mathbf{F}^{\text{C}}\right\rbrace$) in terms of table elements are extracted, \textit{ i.e.,} \textit{geometry}, \textit{appearance} and \textit{content embeddings}. In each collaborative block, the extracted feature embeddings are built as context graphs which are separately applied by the ECE to shape ``intra-modality stream''. Afterwards, the CCS selectively fuses individual contextual information from different modalities as inter-modality interactions maintained in ``inter-modality stream''. Note, we set $\mathbf{M}^{\sim}_{(0)} = \mathbf{F}^{\sim}$ as the initial input of  CCS. The block is stacked $L$ layers to implement the intra-inter modality collaboration in a hierarchical way. To predict the final table structure, the output collaborative graph embeddings from the $l$-th layer of inter-modality stream are sampled as pairs for cells, rows and columns classification.

\subsection{Feature Extraction}
\vspace{-1mm}
In this component, a set of multi-modality features in terms of table elements are extracted from table image, including geometry embeddings $\mathbf{F}^{\text{G}} \in \mathbb{R}^{N \times d}$, appearance embeddings $\mathbf{F}^{\text{A}} \in \mathbb{R}^{N \times d}$ and content embeddings $\mathbf{F}^{\text{C}} \in \mathbb{R}^{N \times d}$. $N$ denotes the number of text segment bounding boxes. 
A more detailed description is given in supplementary material.

\subsection{Collaborative Block}

\paragraph{Ego Context Extractor.}
Now we elaborate on how to extract contextual interactions within each modality of table elements with the help of the Ego Context Extractor~(ECE). Specifically, each extracted modality of features input to the ECE is constructed as individual directed graph $ \mathbf{G}^{\sim} = \left\lbrace \mathcal{V},\mathcal{E} \right\rbrace \in  \left\lbrace\mathbf{G}^{\text{G}},  \mathbf{G}^{\text{A}}, \mathbf{G}^{\text{C}}\right\rbrace$. In each decoupled modality of graph, corresponding embedding of each text segment bounding box is regarded as node  $\mathbf{X} = \left\lbrace \mathbf{x}_1, \mathbf{x}_2, ...,\mathbf{x}_N \right\rbrace \subseteq \mathcal{V}$  which is connected to each other by edges $\mathcal{E} \subseteq \mathcal{V} \times \mathcal{V}$. In the similar spirit with works~\cite{rethinkingGraphs,raja2020table}, we adopt the following asymmetric edge function $h_{\Theta}(\mathbf{x}_i, \mathbf{x}_j) =\mathbf{x}_i\|(\mathbf{x}_i-\mathbf{x}_j) $  to combine graph edge features to each node, which can be denoted as $\mathbf{H}^{\sim}_{\Theta} \in \mathbb{R}^{(N\cdot(N-1)/2) \times d}$ . In the constructed graphs, each node can be either an anchor or one of context of others. In previous works using DGCNN~\cite{rethinkingGraphs,raja2020table}, only local context of each node is selected by \textit{k}-Nearest Neighbors algorithm~(KNN) to be aggregated into node feature. However, the local context is not versatile for representing relationships of all modalities. Besides, the DGCNN-based methods apply CNN to perform local context aggregation. For graph representation, CNN with strong inductive bias (\textit{e.g.,} local behavior) may not be the optimal choice. To tackle the above problems, our proposed ECE instead aggregates information of fully-connected graph for all three modalities via Multi-head Attention (MHA)~\cite{vaswani2017attention} module, which has been verified that it makes few assumptions about inputs and can learn to combine local behavior and global information based on input content~\cite{cordonnier2020relationship}.  

\begin{figure}[htb]
	\begin{center}
				\vspace{-3mm}
		\includegraphics[width=1\linewidth]{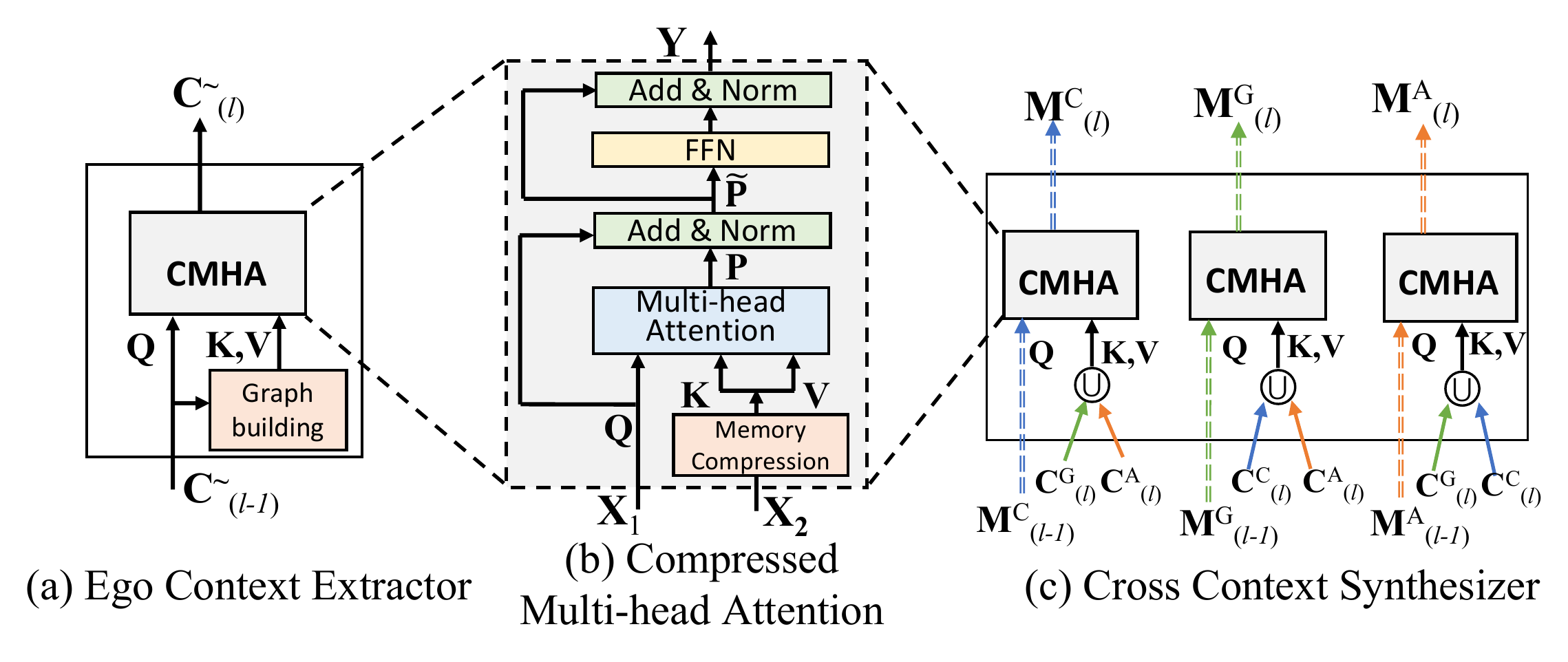}
	\end{center}
	\vspace{-6mm}
	\caption{The proposed Ego Context Extractor and Cross Context Synthesizer Modules in collaborative block. Best viewed in color.} 
	\label{fig:ec}
	\vspace{-1mm}
\end{figure}

More concretely, \textit{l}-th ECE takes intra-modality features $\mathbf{C}^{\sim}_{(\textit{l-1})}$ as queries $\mathbf{Q}$ and the graph edge combined features  $\mathbf{H}^{\sim}_{\Theta}$ as keys $\mathbf{K}$ and values $\mathbf{V}$ as illustrated in Fig.~\ref{fig:ec}(a). Note, for the first layer, we input $\mathbf{F}^{\sim}$ as $\mathbf{C}^{\sim}_{(\textit{0})}$.   However, the main limitation of using MHA is that the amount of input $\mathbf{K}$ and  $\mathbf{V}$ can be very large ( $N\cdot(N-1)/2$ in our case), which is infeasible to be trained. Given $\mathbf{Q}\in \mathbb{R}^{N \times d_q}, \mathbf{K}\in \mathbb{R}^{M \times d_k}, \mathbf{V}\in \mathbb{R}^{M \times d_v}$ and $M= N\cdot(N-1)/2$, the time complexity of the attention operation is $\mathcal{O}(NM)$ and the output is in $N \times d_v$ dimensionalities, of which the number is only relevant to that of $\mathbf{Q}$. Therefore, we can extend the MHA to a more memory-efficient Compressed MHA~(CMHA) by introducing memory compression module which is utilized to reduce image pixel numbers in~\cite{wang2021pyramid}, as depicted in Fig.~\ref{fig:ec}(b). In detail, the compression operation can be implemented as: 
\vspace{-1mm}
\begin{align}
	MC(\mathbf{H}) = Norm(Reshape(\mathbf{x},\epsilon)\mathbf{W}^h),
\end{align}
where $Reshape(\mathbf{H},\epsilon)$ denotes the operation of reshaping input $\mathbf{x} \in \mathbb{R}^{M \times d}$ to  $\widetilde{\mathbf{x}} \in \mathbb{R}^{\epsilon M \times d/\epsilon}$, and $\epsilon \in \left[ 0,1\right] $ is the compression ratio. Through this way, the complexity can be quadratically reduced from $\mathcal{O}(NM)$ to $\mathcal{O}(N\epsilon M)$. In default, we set $\epsilon = N/M$, where $N$ is the number of queries $\mathbf{Q}$. And $Norm(\cdot)$ is the layer normalization. Additionally, we also equip the CMHA with residual connections in our method to make the query information flow unimpeded, which can be defined as:
\vspace{-1mm}
\begin{align}
	\small
	&\mathbf{Y}= Add\&Norm(FFN(\mathbf{\widetilde{P}}),\mathbf{\widetilde{P}}),\\
	&\mathbf{\widetilde{P}}= Add\&Norm(\mathbf{Q},\mathbf{P}),\\
	&\mathbf{P} = MHA(\mathbf{Q},MC(\mathbf{K}), MC(\mathbf{V})),
\end{align}
where ``$FFN(\cdot)$'' is the feed-forward layer and ``$Add\&Norm(\cdot)$'' denotes element-wise addition and layer normalization, which is similar to the work~\cite{vaswani2017attention}. Conclusively, the contextual graph information is baked into graph node as $\mathbf{C}^\sim \in \left\lbrace\mathbf{C}^{\text{G}},  \mathbf{C}^{\text{A}}, \mathbf{C}^{\text{C}}\right\rbrace$ within each modality through the CMHA in our ECE module. 
\vspace{-0.2cm}
\paragraph{Cross Context Synthesizer.}
Once heterogeneous context graph embeddings are obtained, our goals are to fuse them together in a collaborative way and to learn the collaborative patterns between different modalities. Also based on the CMHA, we design the Cross Context Synthesizer (CCS), as is shown in Fig.~\ref{fig:ec}(c). In detail, the CCS has three parallel CMHA modules, and each of them takes one modality as queries while the other two are jointly regarded as keys and values. Take the first branch in Fig.~\ref{fig:ec}(c) for example, the CMHA takes ``content'' modality of context graph embeddings as $\mathbf{Q}$, and the respective outputs of ECE for ``geometry'' and ``appearance'' are input as $\mathbf{K}$ and $\mathbf{V}$. In Fig.~\ref{fig:ec}(c), ``\textcircled{\small{U}}'' denotes the union of two modality sets. For the similar purpose in ECE process, we also follow the similar rule to compress the number of ``memory'' to $N$ which equals to that of $\mathbf{Q}$. Essentially, the query modality explores helpful information from another two modalities.  

\subsection{Table Structure Prediction}\label{sec:tsp}
At the $l$-th layer of collaborative block, the outputs of CCS are to further fused as collaborative graph embeddings, which are denoted as $\mathbf{E} = \left\lbrace \mathbf{{e}}_1, \mathbf{{e}}_2, ..., \mathbf{{e}}_N \right\rbrace \in \mathbb{R}^{N \times d_e} $. Based on the embedings $\mathbf{E}$, our method constructs the $i$-th and $j$-th samples as pairs and concatenate them along channel axis as vectors $\mathbf{U} = \left\lbrace \mathbf{{u}}_{1,1}, \mathbf{{u}}_{1,2}, ...,\mathbf{{u}}_{i,j},..., \mathbf{{u}}_{N,N} \right\rbrace \in \mathbb{R}^{N^2 \times 2d_e}$. Then three groups of FC layers are separately applied for predicting binary-class relations of $\mathbf{U}$, \textit{i.e.,} whether the pair of \textit{i}-th and \textit{j}-th sample is belong to the same row, column or cell, as illustrated in Fig.~\ref{fig:arch}. Each FC group consists of three FC layers with 256 dimensions and a 2-dimension FC with softmax layer. 

\subsection{Training Strategy}
We train our proposed NCGM in an end-to-end way. The whole loss function is defined as $\mathcal{L} = \mathcal{L}_{cell} + \mathcal{L}_{col}+ \mathcal{L}_{row}$, where $\mathcal{L}_{cell}$, $\mathcal{L}_{col}$ or $\mathcal{L}_{row}$ represents cell, column and row relationship losses. For each of them, we adopt the multi-task loss $\mathcal{L}_{\sim} = \lambda_{1}\mathcal{L}_{class} +\lambda_{2}\mathcal{L}_{con}$ to satisfy both the contrastive objective and to predict belonging classes of the output embedding pairs. $\mathcal{L}_{con}$ and $\mathcal{L}_{class}$ are contrastive loss and binary classification loss functions respectively. A more detailed description is given in supplementary material.

\section{Experiments}\label{sec:experiments}
\subsection{Datasets and Evaluation Protocol} \label{sec:data}

\paragraph{Datasets.}
We perform extensive experiments on various benchmark datasets. Among, 
ICDAR-2013~\cite{gobel2013icdar}, ICDAR-2019~\cite{gao2019icdar}, {WTW\cite{long2021parsing}}, UNLV~\cite{shahab2010open}, SciTSR~\cite{chi2019complicated} and SciTSR-COMP~\cite{chi2019complicated} are employed for physical structure recognition, while TableBank~\cite{li2020tablebank} and PubTabNet~\cite{end2end1} are adopted for evaluating logical structure recognition performance. It should be noted that there is no training set in ICDAR-2013 and UNLV datasets, so we extend the two datasets to the partial versions, which is similar to TabStruct-Net~ \cite{raja2020table}. A more detailed description about public benchmarks is given in supplementary material.

To further investigate the capacity of our proposed method under more challenging scenarios, we expand ``SciTSR-COMP'' dataset to ``SciTSR-COMP-A'' by applying two kinds of distortion algorithms. A more detailed description is given in supplementary material.

\vspace{-4mm}
\paragraph{Evaluation settings.}
Several existing works are only applicable to table images alone,
while others utilize additional information including text segment/cell bounding boxes or text contents.
To compare in a unified protocol, we follow two different experimental setups in~\cite{raja2020table}: (a) Setup-A where only table image is taken as input without additional information and (b) Setup-B where table image along with additional features such as cell/text segment bounding boxes and text contents.
For a fair comparison, {we also incorporate the result boxes of detection in FLAG-Net~\cite{liu2021show}} and the OCR results of Tesseract~\cite{smith2007overview} as inputs in Setup-A. 

\vspace{-4mm}
\paragraph{Evaluation protocol.}
We employ precision, recall and F1-score~\cite{gobel2012methodology} as protocol to evaluate the performance of our model for recognizing table physical structure including vertical and horizontal relations.
For the recognition of table logical structure, BLEU score~\cite{papineni2002bleu} used in \cite{li2020tablebank} and Tree-Edit-Distance-based Similarity (TEDS) proposed in~\cite{end2end1} are exploited.

\subsection{Implementation Details}\label{sec:details}
The framework is built on Pytorch~\cite{paszke2019pytorch}.
We scale the input table images to a fixed size $512\times512$ to introduce scale invariance.
In default, the layer number of collaborative blocks is set to 3 and the hidden size $d$ is set to 64. Further, we set $h = 8, d_m = 64, d_k = d_v = 8$ for both Ego Context Extractor~(ECE) and Cross Context Synthesizer~(CCS) of each collaborative block. 
During training, the learning rate is initialized as $1\mathrm{e}{-4}$ and divided by 10 when the loss stops decreasing. For the training loss, we empirically set all weight parameters $\lambda_{1}=\lambda_{2}=1$.
For all experiments, the models are pre-trained on SciTSR for 10 epochs, and then fine-tuned on different benchmarks for 50 epochs, 
which is conducted on the platform with one Nvidia Tesla V100 GPU and 32 GB memory.

\subsection{Comparison with State-of-the-arts}
\begin{figure}[htb]
	\centering
	\subfloat[Sample result of {FLAG-Net} on SciTSR-COMP-A dataset.]{
		\includegraphics[width=0.47\columnwidth, height=2cm]{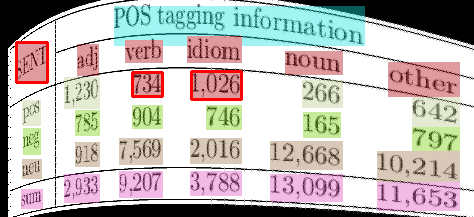}
		\hspace{2mm}
		\includegraphics[width=0.47\columnwidth, height=2cm]{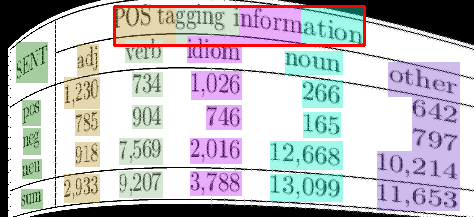}
	}

	\subfloat[Sample result of NCGM on SciTSR-COMP-A dataset.]{
		\includegraphics[width=0.47\columnwidth, height=2cm]{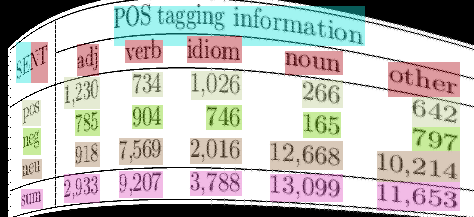}
		\hspace{2mm}
		\includegraphics[width=0.47\columnwidth, height=2cm]{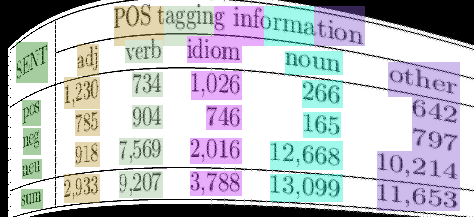}
	}
	\vspace{-0.2cm}
	\caption{Visualization of physical relationships of distorted table between {FLAG-Net} and NCGM. The first and second column indicate the predictions of rows and columns respectively. The boxes belonging to the same relationships are filled in the same colors. The boundaries of the text segment boxes with misrecognized relationships are marked in red lines. Our NCGM shows better tolerance for the challenging scenarios compared with {FLAG-Net}.}
	\label{fig:result} 
\end{figure}

\paragraph{Results of physical structure recognition.}
As is shown in Tab. \ref{Comparison_physical}, our NCGM outperforms most of previous methods on different datasets for physical structure recognition. {Compared with the strong baseline FLAG-Net~\cite{liu2021show}, NCGM increases average F1-score on all datasets by round 2\% under both Setup-A settings and Setup-B settings. When processing table images with complex distortions (``SciTSR-COMP-A''), it is worth mentioning that our NCGM can achieve about 11\% and 12\% higher F1-scores under Setup-A and Setup-B than the second-best FLAG-Net~\cite{liu2021show} without using distorted images as training data. If taking distorted data as training set, the performance of NCGM still can surpass it round 7\% and 9\% under both settings respectively.} We also visualize row and column physical relationships of distorted table in Fig.~\ref{fig:result}. Note, the different color blocks in it merely visualize the belonging relationship rather than dividing the entire box. Taking right column of Fig.~\ref{fig:result} for example, ``POS tagging information'' is one whole text segment bounding box. In logical, one can observe that ``POS tagging information'' box spans across five columns of word bounding boxes below it in column dimension. Therefore, the five columns attribute their respective colors to the ``POS tagging information'' box. By comparison, our method correctly recognizes both relationships while the {FLAG-Net} performs unsatisfactorily under distorted table scenes.

\vspace{-4mm}
\paragraph{Results of logical structure recognition.}
In order to evaluate our model on logical structure recognition task benchmarks, \textit{i.e.,} TableBank and PubTabNet, we perform lightweight post-processing (see supplementary material) on the NCGM's output results of row/column relationships to convert them to the HTML representation.
Tab. \ref{Comparison_logical1} presents that our method achieves significant improvement compared with other methods for logical structure recognition task.

\paragraph{Computational complexity.} A more detailed description is given in supplementary material.
\begin{table}[htb!]
	\centering
	\small
	\setlength{\tabcolsep}{0.6mm}
	\scalebox{1} {
		\begin{tabular}{l|l|ccc|ccc}
			\toprule[1.5pt]
			\multicolumn{8}{c}{\textbf{ICDAR-2013-P}} \\
			\hline
			\multirow{2}{*}{Method} & \multirow{2}{*}{ \tabincell{l}{Train Dataset}} & \multicolumn{3}{c|}{Setup-A} & \multicolumn{3}{c}{Setup-B} \\
			\cline{3-8}
			&& P & R & F1 & P & R & F1 \\
			\hline
			DGCNN~\cite{rethinkingGraphs} &  \tabincell{l}{Sci. + IC13-P} & - &-  &-& 98.6 &99.0 &98.8   \\
			TabStr.~\cite{raja2020table} &   \tabincell{l}{Sci. + IC13-P} & 93.0 &90.8 &91.9 &99.1 &99.3 &99.2 \\
			GTE~\cite{zheng2021global} &  \tabincell{l}{Pub. + IC13-P} & 94.4 &92.7  &93.5&- & -  & -  \\
			LGPMA~\cite{qiao2021lgpma} &  \tabincell{l}{Sci. + IC13-P} & 96.7 &99.1 &97.9& - & -  & - \\
			{C-CTRNet~\cite{long2021parsing}} & WTW + IC19 & 95.5 &88.3 & 91.7&  - & -  & - \\
			{FLAG-Net~\cite{liu2021show}} & \tabincell{l}{Sci. + IC13-P} & 97.9&\textbf{99.3}&98.6 &99.2 &99.5 &99.3\\
			\hline
			\textbf{NCGM} & \tabincell{l}{Sci. + IC13-P} & \textbf{98.4} & \textbf{99.3}   & \textbf{98.8}  & \textbf{99.3}  & \textbf{99.9}  & \textbf{99.6}  \\
			\midrule[1.5pt]	
			\multicolumn{8}{c}{\textbf{ICDAR-2019}} \\
			\hline
			DGCNN~\cite{rethinkingGraphs} & \tabincell{l}{Sci. + IC19} & 80.3& 77.8 &79.0&- &-  &-   \\
			TabStr.~\cite{raja2020table} & \tabincell{l}{Sci. + IC19} &  82.2 & 78.7 &80.4 &97.5 &95.8 &96.6 \\
			{C-CTRNet~\cite{long2021parsing}}&WTW& -&-&80.8& - &-  &-   \\
			{FLAG-Net~\cite{liu2021show}}& \tabincell{l}{Sci. + IC19} & \textbf{85.2}&83.8&84.5&96.1 &96.3  &96.2 \\
			\hline
			\textbf{NCGM} & \tabincell{l}{Sci. + IC19} &  {84.6} &	\textbf{86.1}&	\textbf{85.3} &\textbf{98.9} &\textbf{98.8} &\textbf{98.8}  \\
			\midrule[1.5pt]
			\multicolumn{8}{c}{\textbf{WTW}} \\
			\hline
			{C-CTRNet~\cite{long2021parsing}} & \tabincell{l}{WTW} & 93.3 &91.5 &92.4&- & -  & - \\
			FLAG-Net~\cite{liu2021show} &WTW& 91.6&89.5&90.5& 93.2 &91.7  &92.4  \\
			\hline
			\textbf{NCGM}& \tabincell{l}{WTW} & \textbf{93.7} & \textbf{94.6} & \textbf{94.1} &\textbf{95.8} & \textbf{96.4} & \textbf{96.1}  \\
			\midrule[1.5pt]
			\multicolumn{8}{c}{\textbf{UNLV-P}} \\
			\hline
			DGCNN~\cite{rethinkingGraphs} & \tabincell{l}{Sci. + UNLV-P} & - & -  & -& 92.1 &89.8 &90.9 \\
			TabStr.~\cite{raja2020table} &  \tabincell{l}{Sci. + UNLV-P} & 84.9 &82.8 &83.9 &99.2 &99.4 &99.3 \\
			{FLAG-Net~\cite{liu2021show}} & \tabincell{l}{Sci. + UNLV-P} &\textbf{89.2}&87.3&88.2&98.9 &99.5  &99.2\\
			\hline
			\textbf{NCGM}& \tabincell{l}{Sci. + UNLV-P} & {88.9} &\textbf{88.2} &\textbf{88.5} &\textbf{99.8} &\textbf{99.8} &\textbf{99.8}  \\
			\midrule[1.5pt]
			\multicolumn{8}{c}{\textbf{SciTSR}} \\
			\hline
			DGCNN~\cite{rethinkingGraphs} & Sci. & - & -  & -& 97.0 &98.1 &97.6 \\
			TabStr.~\cite{raja2020table} &  Sci. & 92.7 &91.3 &92.0 &98.9 & 99.3 &99.1 \\
			LGPMA~\cite{qiao2021lgpma} & Sci. & 98.2 &99.3 &98.8&  - &-  &- \\
			{FLAG-Net~\cite{liu2021show}} & Sci.& \textbf{99.7} &99.3 &99.5&\textbf{99.8} &99.5 &99.6\\
			\hline
			\textbf{NCGM}& Sci.& \textbf{99.7} &\textbf{99.6} &\textbf{99.6} &{99.7} & \textbf{99.8} &\textbf{99.7}  \\
			\midrule[1.5pt]
			\multicolumn{8}{c}{\textbf{SciTSR-COMP}} \\
			\hline
			DGCNN~\cite{rethinkingGraphs} & Sci. & - &-  &-& 96.3 &97.4 &96.9 \\
			TabStr.~\cite{raja2020table} &  Sci. & 90.9 &88.2 &89.5 &98.1 &98.7 &98.4 \\
			LGPMA~\cite{qiao2021lgpma} & Sci. & 97.3 & 98.7 &98.0&- &-  &- \\
			{FLAG-Net~\cite{liu2021show}} & Sci. & 98.4 &98.6 &98.5&98.6 &99.0 &98.8 \\
			\hline
			\textbf{NCGM}& Sci.& \textbf{98.7} & \textbf{98.9}&\textbf{98.8} & \textbf{98.8} & \textbf{99.3} & \textbf{99.0}  \\
			\midrule[1.5pt]
			\multicolumn{8}{c}{\textbf{SciTSR-COMP-A}} \\
			\hline

			{FLAG-Net~\cite{liu2021show}}& Sci. & 70.7 & 66.2 & 68.4  &83.3 &81.0  &82.1  \\
			{FLAG-Net~\cite{liu2021show}}& Sci. + Sci.-C-A & 82.5 & 83.0 & 82.7  &88.8 &87.5  &88.1  \\
			\hline
			\textbf{NCGM}& Sci.& \textbf{79.6} & \textbf{78.9}  & \textbf{79.2}  & \textbf{93.3} & \textbf{94.8} & \textbf{94.0}  \\
			\textbf{NCGM}& Sci. + Sci.-C-A& \textbf{88.4} & \textbf{90.7}  & \textbf{89.5}  & \textbf{97.2} & \textbf{97.5} & \textbf{97.3}  \\
			\bottomrule[1.5pt]
		\end{tabular}
	}
	\caption{Comparison results of physical structure recognition on ICDAR-2013-P, ICDAR-2019, {WTW}, UNLV-P, SciTSR, SciTSR-COMP and SciTSR-COMP-A dataset. ``-P'' means partial dataset and ``-A'' represents augmented dataset by distortion. ``P'', ``R'' and ``F1'' stand for ``Precision'', ``Recall'' and ``F1-score'' respectively. ``TabStr.'' and ``C-CTRNet'' denote  ``TabStruct-Net'' and ``Cycle-CenterNet'' individually.}
	\label{Comparison_physical}
\end{table}

\begin{table}[htb]
	\small
	\centering
	\setlength{\tabcolsep}{5mm}
	\begin{tabular}{l|l|c}
		\toprule[1.5pt]
		\multicolumn{3}{c}{\textbf{TableBank}} \\
		\hline
		\multirow{2}{*}{Method} & \multirow{2}{*}{Train Dataset} & Setup-A\\
		\cline{3-3}
		&& BLEU \\
		\hline
		Image-to-Text~\cite{li2020tablebank} & TableBank & 73.8 \\
		TabStruct-Net~\cite{raja2020table}  & SciTSR & 91.6\\
		{FLAG-Net~\cite{liu2021show}} & SciTSR & 93.9 \\
		\hline
		\textbf{NCGM} & SciTSR& \textbf{94.6}\\
		\midrule[1.5pt]
		\multicolumn{3}{c}{\textbf{PubTabNet}} \\
		\hline
		\multirow{2}{*}{Method} & \multirow{2}{*}{Train Dataset} & Setup-A\\
		\cline{3-3}
		&& TEDS \\
		\hline
		EDD~\cite{end2end1} & PubTabNet & 88.3  \\
		TabStruct-Net~\cite{raja2020table}  &  SciTSR & 90.1 \\
		GTE~\cite{zheng2021global} & PubTabNet & 93.0 \\
		LGPMA~\cite{qiao2021lgpma} & PubTabNet & 94.6  \\
		{FLAG-Net~\cite{liu2021show}} & SciTSR & 95.1 \\
		\hline
		\textbf{NCGM} &  SciTSR& \textbf{95.4}\\
		\bottomrule[1.5pt]
	\end{tabular}
	\caption{Comparison results of logical structure recognition on TableBank and PubTabNet datasets.}
	\label{Comparison_logical1}
\end{table}

\subsection{Ablation Study}\label{sec:ablation}
In this subsection, we perform several analytic experiments under Setup-B settings on SciTSR-COMP benchmark to investigate the contributions of intra-modality and inter-modality interactions in our proposed NCGM. 
\paragraph{Effect of intra-modality interactions.}
For intra-modality interactions, Tab.~\ref{Ablation} compares the effectiveness of various extractors, including DGCNN~\cite{rethinkingGraphs} and Transformer~\cite{vaswani2017attention}, with ECE in our method. ``Mixed'' means all modality features are \textbf{early-fused} by concatenation as the input and ``Individual'' denotes each modality is input into context extractor separately. 
Tab.~\ref{Ablation} shows ECE can achieve the best performance when taking either mixed features or individual features as input while ``Transformer'' performs the worst. For ``DGCNN'', it only aggregates information from top K similar nodes of each node instead of all ones. 
Compared with ``DGCNN'', although ``Transformer'' can deploy the global information of nodes, it ignores the directed edge effects between nodes. Encouragingly, our CMHA-based ECE can not only consider the directed relationships between nodes, but also extract the context information from all nodes.
Additionally, we can also observe that individual features can yield better results than the mixed ones, which proves that decoupling the individual modality from each other is indeed a more preferable way to solve the Hetero-TSR problem. 

\begin{table}[htb]
	\small
	\setlength{\tabcolsep}{0.8mm}
	\centering
	\scalebox{1} {
		\begin{tabular}{c|cc|ccc|cc|ccc}
			\toprule[1.5pt]
			\multirow{2}{*}{\tabincell{c}{Fusion\\ Method}} & \multicolumn{2}{c|}{{Input}} & \multicolumn{3}{c|}{{Intra.}} & \multicolumn{2}{c|}{{Inter.}} & \multicolumn{3}{c}{Setup-B}\\
			\cline{2-11}
			&Mix.&Ind.&DG.&Tr.&ECE&Con.&CCS& P & R& F1 \\
			\hline
			\multirow{3}{*}{\tabincell{c}{Early\\ Fusion}} &\cmark & \xmark & \cmark & \xmark& \xmark & \xmark & \xmark & 96.3 &97.4 &96.8  \\
			&\cmark & \xmark & \xmark & \cmark& \xmark & \xmark & \xmark & 95.1  & 95.6& 95.3  \\
			&\cmark & \xmark & \xmark & \xmark& \cmark & \xmark & \xmark & 97.8 & 98.3 & 98.0  \\
			\hline
			\multirow{3}{*}{\tabincell{c}{Late\\ Fusion}} &\xmark & \cmark & \cmark & \xmark& \xmark & \cmark & \xmark & 96.9 & 98.2 & 97.5  \\
			&\xmark & \cmark & \xmark & \cmark& \xmark & \cmark & \xmark & 94.9  & 96.1& 95.5  \\
			&\xmark & \cmark & \xmark & \xmark& \cmark & \cmark & \xmark &  98.4 & 98.2 & 98.3   \\
			\hline
			\textbf{NCGM} &\xmark & \cmark & \xmark & \xmark& \cmark & \xmark & \cmark & \textbf{98.8} & \textbf{99.3} & \textbf{99.0} \\
			\bottomrule[1.5pt]
		\end{tabular}
	}
	\caption{Ablation studies of NCGM on SciTSR-COMP dataset. ``Intra.'' and ``Inter.'' stand for intra-modality interactions and inter-modality interactions respectively. ``Mix.'' and ``Ind.'' are short for ``Mixed'' and ``Individual''. ``DG.'' and ``Tr.'' denote ``DGCNN'' and ``Transformer''. ``Con.'' represents ``Concatenation''.}
	\label{Ablation}
	\vspace{-0.5cm}
\end{table}

\paragraph{Effect of inter-modality interactions.}
We compare the proposed CCS with the ``Concatenation'' operation of multi-modal features in Tab.~\ref{Ablation}. It can be observed that CCS improves the accuracy of predicting adjacency relationship compared with directly \textbf{late-fused} multiple model features via concatenation.
This confirms the benefits of CCS that enables one modality to positively collaborate with the others, and can capture the complex implicit modality relationships.
Moreover, it also proves that the CCS module combined with ECE can further boost the performance.

\subsection{Further Analysis on Collaborative Block}
\begin{figure*}[htb]
	\includegraphics[width=1\linewidth,height=0.6\linewidth]{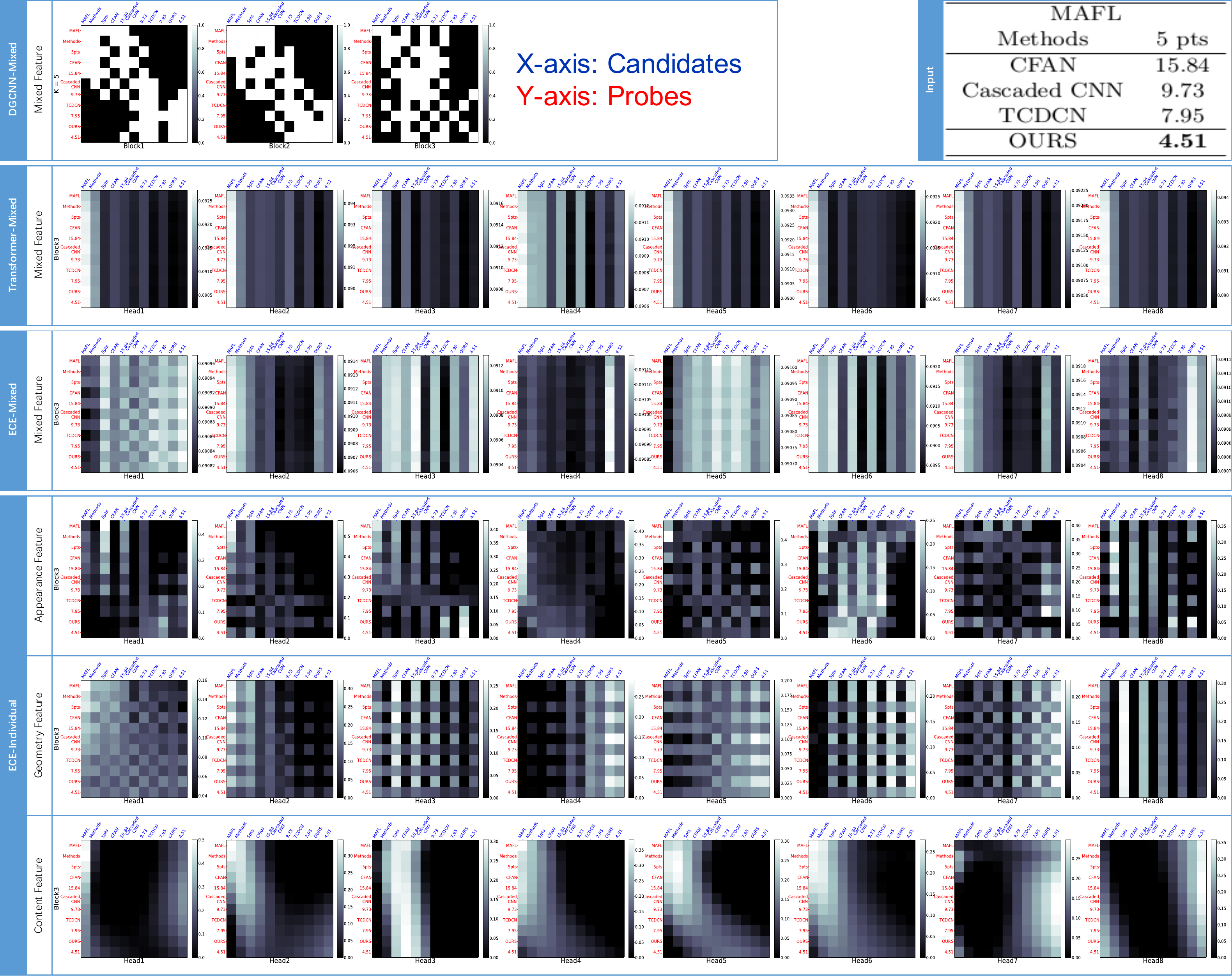}
	\caption{ Visualization of the heat-maps generated by DGCNN and multi-head attention maps from the Transformer and ECE. Y-axis (red) and X-axis (blue)  are ``probes'' and ``candidates'' respectively. For ECE, probes are graph node features and candidates are edge combined features. For Transformer and DGCNN, probes and candidates are both non-graph features.  The heat-maps of DGCNN show a local hard selection way in terms of context. And Transformer yields attention maps lacking sparsity and diversity. In contrast, ECE-Mixed presents more diversified attention maps and ECE-Individual extracts interactions in global or local pattern conditioned on different features. Best viewed in color.
	} 
	\label{fig:heatmap}
		\vspace{-0.4cm}
\end{figure*}

\paragraph{What does ECE learn from the intra-modality?}
As suggested by recent works~\cite{malaviya2018sparse,raganato2018analysis,marevcek2018extracting} on interpreting attention mechanism, separate attention heads may learn to look for various relationships between inputs and introducing more sparsity and diversity for attention may improve performance and interpretability.
To explore the intra-modality interactions learned by ECE in collaborative block, we in Fig. \ref{fig:heatmap} visualize the multi-head attention maps from last blocks of ECE. For comparison, we also visualize the multi-head self-attention maps from the last blocks of ``Transformer-Mixed''~\cite{vaswani2017attention} and KNN (K = 5) selection heat-maps of all layers in DGCNN~\cite{rethinkingGraphs}, where a lighter color indicates a closer relationship.
The KNN results of DGCNN show that the feature aggregation of one node only pays attention to the top K similar features of other nodes instead of all the nodes, and relies on the choice of K. The attention maps of Transformer-Mixed present equilibrium status, which lacks sparsity and diversity. Comparatively, our ``ECE-Mixed'' taking mixed features presents more diversified attention maps in eight heads, which indicates ECE can more effectively capture context information. Moreover, the attention maps generated by ``ECE-Individual'' show different intriguing focus patterns for different features. Specifically, ECE prefers to extract interactions for appearance and geometry features in global scope while content features bring more local focus patterns.

\begin{figure}[htb]
	\centering
	\includegraphics[width=0.94\columnwidth, height=4.7cm]{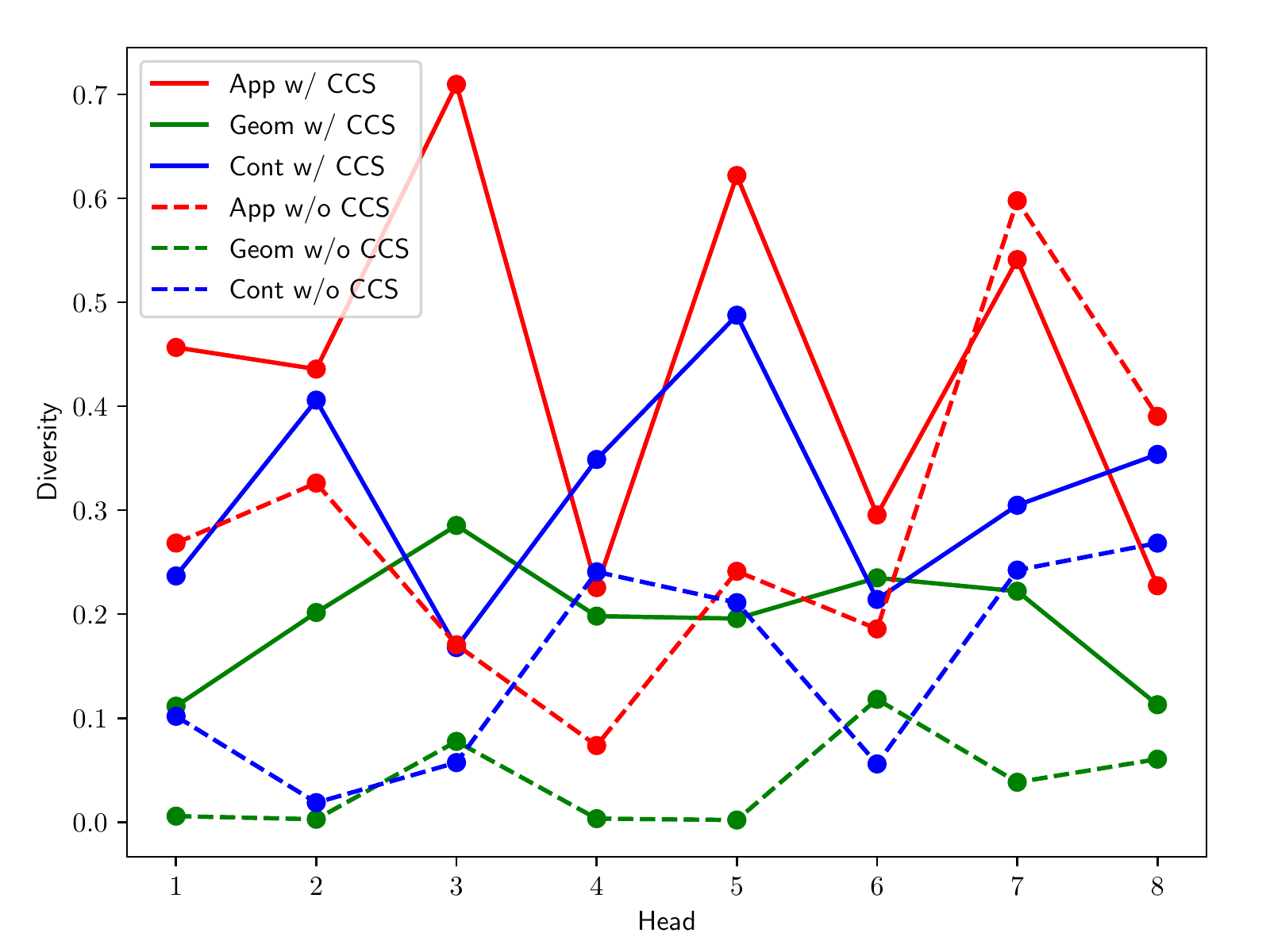}\\
	\vspace{-0.2cm}
	\caption{Diversities of attention maps for different modalities with or without CCS. Solid lines~($\sim$ w/ CCS) represent the diversity distributions of attention in CCS when one modality features are regarded as queries and others as keys/values. Dashed lines~($\sim$ w/o CCS) present diversity of attention weights in ECE for each modality.} 
	\label{fig:ccs} 
\end{figure}

\vspace{-0.7cm}
\paragraph{How do different modalities collaborate with each other?}
To investigate the working pattern of CCS, we adopt Jensen-Shannon Divergence~\cite{correia2019adaptively}~(see supplementary materials) to measure the average diversity of attention map in CCS when the model also takes input table image shown in Fig.~\ref{fig:heatmap}. As shown in Fig.~\ref{fig:ccs}, solid lines~($\sim$ w/ CCS) represent the diversity distributions when one modality features are regarded as queries and others as keys/values. After removing CCS, diversity of attention weights in ECE for each modality is also presented by dashed lines~($\sim$ w/o CCS). For those with CCS, the higher value indicates the query modality is in a closer collaboration with the other modalities. Particularly, appearance modality has the strongest collaborative relationship with others while geometric one requires the least collaboration. By comparison, the diversities of attention weights in ECE also follow a similar trend, but with lower values on average. 

\begin{figure}[htb]
	\centering
	\includegraphics[width=0.9\columnwidth, height=4.6cm]{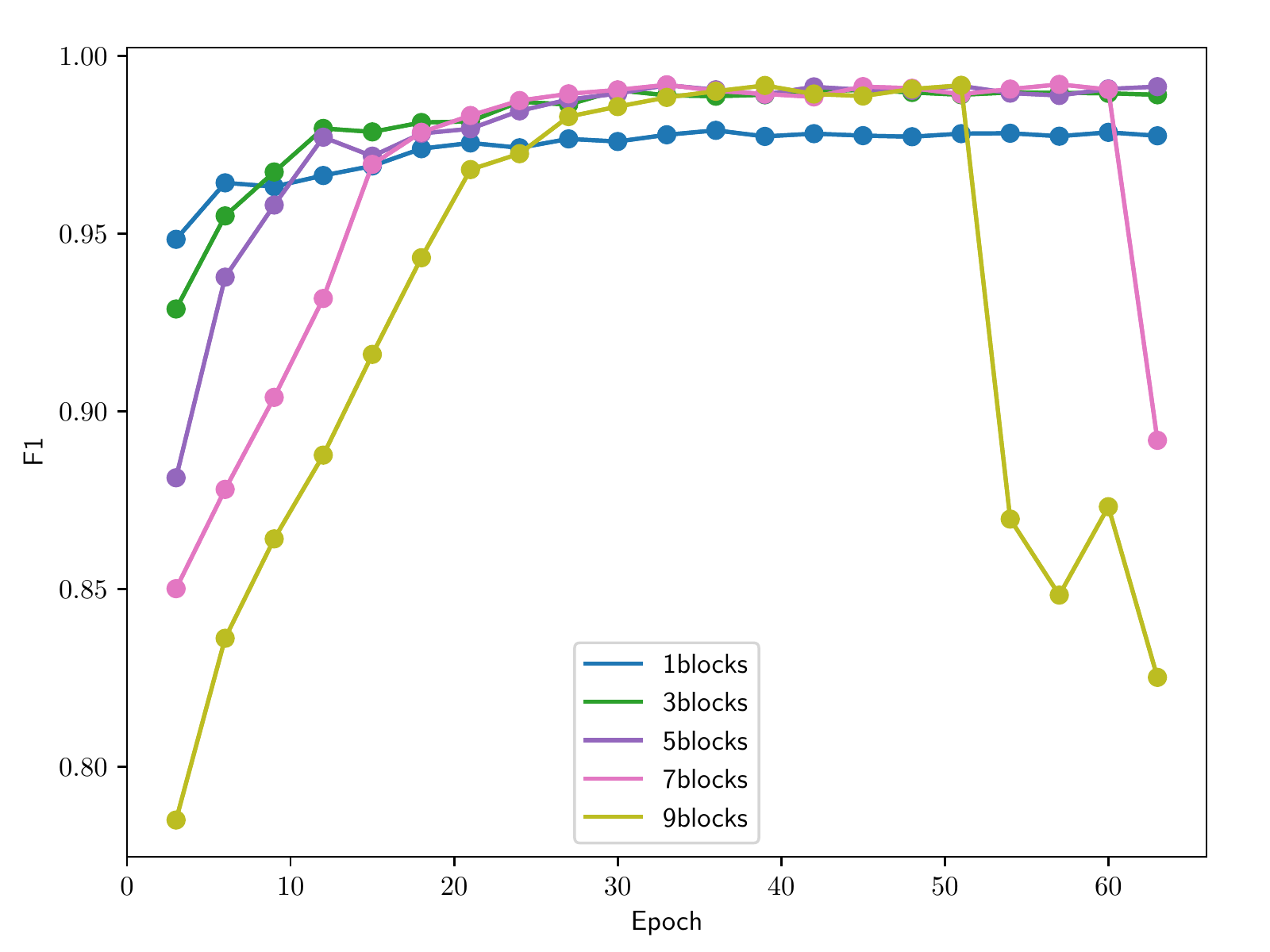}\\
	\vspace{-0.2cm}
	\caption{The relationship between block number of NCGM and F1-score on SciTSR-COMP dataset.} 
	\label{fig:block} 
	\vspace{-0.6cm}
\end{figure}

\paragraph{The more collaborative blocks, the better performance?}
To further explore the effect of the collaborative block number on the NCGM performance, we conduct a set of experiments setting block numbers from 1 to 9, respectively.
It can be seen from Fig.~\ref{fig:block} that it is a trade-off problem. Small block number can render faster convergence to the model. As the number increases, the performance keeps improving until block number increases to 5, but the convergence speed of the network keeps slowing down. In particular, we observe that the F1-score decreases sharply when NCGM with more than 7 blocks is trained over round 50 epochs, which indicates more blocks are easier to cause model training collapse problem. Based on the above observation, we set it to 3 as default number.

\section{Conclusion and Limitation}\label{sec:conclusion}

We present a novel graph-based method for heterogeneous table structure recognition through learning intra-inter modality collaboration. Extensive experiments on public benchmarks demonstrate its superiority over state-of-the-art methods, especially under challenge scenarios. 
There still exist two limitations that can be improved in future work. The first one is the inevitable problem of computational complexity increase when introducing multiple modalities and decoupled processing. The second one lies in the fact that NCGM with deeper blocks is easier to suffer from the training collapse problem. We may introduce more inductive bias into the attention model to tackle it.

\begin{appendices}
	\section{Feature Extraction}
	\subsection{Multi-modality Features}
	\paragraph{Geometry embedding.} We derive the geometry feature of each text segment bounding box as 
	$\left({\frac{x}{W}}, {\frac{y}{H}}, {\frac{w}{W}}, {\frac{h}{H}} \right)^\top$, where $W$ and $H$ are the width and height of the table image.  $(x,y)$ represents the center point of the box while height $h$ and width $w$ correspond to its short side and long side respectively. Then a $d$-dimension Fully-Connected (FC) layer is applied on the above vectors to obtain the geometry embeddings $\mathbf{F}^{\text{G}} = \left\lbrace \mathbf{g}_1, \mathbf{g}_2, ...,\mathbf{g}_N \right\rbrace \in \mathbb{R}^{N \times d}$.
	
	\paragraph{Appearance embedding.} 
	We employ ResNet18-based CNN~\cite{he2016deep} as backbone to extract whole table image feature. In detail, the backbone consists of \textit{conv1} to \textit{conv2\_2} of ResNet18 followed by three convolutional layers of size $3\times3\times64$. Hereafter, the output of backbone is applied by the RoI Align~\cite{he2017mask} in terms of text segment bounding boxes. After passing a FC layer with $d$ dimensions,  appearance embeddings $\mathbf{F}^{\text{A}} = \left\lbrace \mathbf{f}_1, \mathbf{f}_2, ...,\mathbf{f}_N \right\rbrace \in \mathbb{R}^{N \times d}$ are obtained. 
	
	\paragraph{Content embedding.} 
	First, we embed corresponding text of each text segment bounding box in distributional space via word2vec~\cite{church2017word2vec}. Then, one convolutional layer with $7 \times 1 \times d $ kernel size and 1 stride is applied to model text sequential feature as content feature embeddings $\mathbf{F}^{\text{C}} = \left\lbrace \mathbf{t}_1, \mathbf{t}_2, ...,\mathbf{t}_N \right\rbrace \in \mathbb{R}^{N \times d}$. 
	
	\subsection{Ablation Study of Mutil-modalities}
	As shown in Tab.~\ref{Ablation_modality}, we observe that among the three modalities, ``G'' plays a dominant role, followed by ``A'', and finally ``C''. The proposed model leveraging all three modalities can achieve impressive progress under all evaluation metrics. In addition, we also explore the attention weights of individual modality. That is, the attention weights of ``A'' and ``G''  tend to be grid-like, indicating that the model focuses on the spatial position of the row or column in global range. And the attention weights of ``C'' are inclined to emphasize on local successive segment bounding boxes. To sum up, the inductive biases of different modalities are of large disparency.
	
	\begin{table}[htp]
		\centering
		\setlength{\tabcolsep}{4.6mm}
		\scalebox{1} {
			\begin{tabular}{ccc|ccc}
				\toprule[1.5pt]
				\multicolumn{3}{c|}{Input Modality} & \multicolumn{3}{c}{Setup-B} \\
				\hline
				A & G & C & P & R & F1  \\
				\hline
				\cmark & \xmark & \xmark  & 89.8&47.9&62.5     \\
				\xmark & \cmark & \xmark & 97.9&97.7&97.8    \\
				\xmark & \xmark & \cmark & 70.5&39.0&50.2    \\
				\cmark & \cmark & \xmark & 98.6&98.3&98.4    \\
				\xmark & \cmark & \cmark & 98.0&95.0&96.5    \\
				\cmark & \xmark & \cmark & 87.6&89.3&88.4    \\
				\cmark & \cmark & \cmark & \textbf{98.8} &	\textbf{99.3} &	\textbf{99.0}  \\
				\bottomrule[1.5pt]
			\end{tabular}
		}
		\caption{Ablation studies of multi-modalities on SciTSR-COMP dataset. ``A'', ``G'' and ``C'' stand for ``appearance'', ``geometry'' and ``content'' modality respectively.}
		\label{Ablation_modality}
	\end{table}
	
	\section{Multi-head Attention}
	We build the core collaborative block of our method upon Multi-head Attention (MHA)~\cite{vaswani2017attention} module. Here, we briefly introduce it as preliminary knowledge. Given queries $\mathbf{Q}$, keys $\mathbf{K}$ and values $\mathbf{V}$, MHA is defined as:
	\begin{align*}
		\small
		\centering
		&MultiHead(\mathbf{Q,K,V}) = Concat(\mathbf{H}_1,\mathbf{H}_2,...,\mathbf{H}_h)\mathbf{W^*},\\
		& \mathbf{H}_i = Attention(\mathbf{QW}_i^Q,\mathbf{KW}_i^K,\mathbf{VW}_i^V ),  i \in \{1,2,..., h\},\\
		&Attention(\mathbf{Q,K,V}) = softmax(\frac{\mathbf{Q}\mathbf{K}^\top}{\sqrt{d_k}})\mathbf{V},\label{eq:att}
	\end{align*}
	where $d_k$ is the dimension of keys while $h$ is the head number. $\mathbf{W}_i^Q \in \mathbb{R}^{d_m \times d_k},\mathbf{W}_i^K \in \mathbb{R}^{d_m \times d_k},\mathbf{W}_i^V \in \mathbb{R}^{d_m \times d_v}$ and $\mathbf{W}_i^* \in \mathbb{R}^{hd_v \times d_m}$ are projection matrices separately. 
	Essentially, the attention process can be regarded as ``memory accessing'' procedure. 
	
	\section{Training Strategy}

	\subsection{Design of Loss Function}
	The binary classification loss is widely applied in previous graph-based works of table structure recognition (TSR). Particularly, we train our proposed  Neural Collaborative Graph Machines (NCGM) in an end-to-end way to satisfy both the contrastive objective and to predict belonging classes of the output embedding pairs. Given a pair of collaborative graph embeddings~($\{\mathbf{e}_{(a)}, \mathbf{e}_{(b)}\}$) and corresponding concatenated vector $\mathbf{u}_{(a,b)}$, we define the multi-task loss function as:
	\begin{align*}
		\small
		&\mathcal{L} = \mathcal{L}_{cell} + \mathcal{L}_{col}+ \mathcal{L}_{row},\\
		&\mathcal{L}_{\sim} = \lambda_{1}\mathcal{L}_{class} +\lambda_{2}\mathcal{L}_{con},\\
		&\mathcal{L}_{con} = \left \| \mathbf{e}_{(a)} - \mathbf{e}^+_{(b)}  \right \|_{2}^{2} +\mathrm{max}\left\{0, \alpha-  \left \| \mathbf{e}_{(a)} - \mathbf{e}^-_{(b)}  \right \|_{2}^{2} \right\},\\
		&\mathcal{L}_{class} = -\textup{log}(P(z=c|\mathbf{u}_{(a,b)})),\\
		&P(z=c|\mathbf{u}_{(a,b)})=\frac{\exp(S_{c}\mathbf{u}_{(a,b)})}{\sum_{k}\exp(S_{k} \mathbf{u}_{(a,b)})}, c \in \{0,1\},
	\end{align*} 
	where $\mathcal{L}_{\sim}$ represents $\mathcal{L}_{cell}$, $\mathcal{L}_{col}$ or $\mathcal{L}_{row}$, corresponding to cell, column and row relationship loss. $\mathcal{L}_{con}$ is contrastive loss in which $\mathbf{e}^+_{(b)}$ and $\mathbf{e}^-_{(b)}$ are the positive and negative pair of $\mathbf{e}_{(a)}$ respectively. The margin parameter $\alpha$ is set to 1. Correspondingly, $\mathcal{L}_{class}$ is the standard softmax loss in terms of $\mathbf{u}_{(a,b)}$. $z$ is the predicted class for the input pairs, and $S$ is the weight matrix used in the softmax function, and  $S_{c}$ and $S_{k} $ represent the $c$-th and $k$-th column of it, respectively. $c=1$ denotes the concatenated pairs belong to the same cell/column/row, and otherwise $c=0$. They are combined by weight parameters $\lambda_{1}$ and $\lambda_{2}$. Considering memory efficiency, we also introduce Monte Carlo sampling for constructing collaborative graph embedding pairs in the training phase, which is similar to~\cite{rethinkingGraphs}. For inference, the sampling is not performed and we construct all collaborative graph embeddings as pairs.

	\begin{algorithm}[htp]
		\setstretch{0.95}
		\KwInput{$\mathbf{T}, ~\mathbf{GT}_{\sim}$\tcp*{$\mathbf{T}$ denotes input table elements. $\mathbf{GT}_{\sim}$ ~($\mathbf{GT}_\sim \in \left\lbrace\mathbf{GT}_{\textit{cell}},  \mathbf{GT}_{\textit{row}}, \mathbf{GT}_{\textit{col}}\right\rbrace$) represents the Ground Truth of different relationships.}}
		\KwOutput{$\mathbf{F}^{pred}_{\sim}$}
		
		\SetKwFunction{FCMHA}{CMHA}
		\SetKwFunction{FECE}{ECE}
		\SetKwFunction{FCCS}{CCS}
		\SetKwFunction{FMain}{Main}
		
		\tcc{Extract features by Compressed Multi-head Attention.}
		\Fn{\FCMHA{$\mathbf{Q},~\mathbf{K},~\mathbf{V}$}}{
			$\mathbf{Y} ~\leftarrow~ MHA(\mathbf{Q}, ~MC(\mathbf{K}), ~MC(\mathbf{V}))$
			
			\KwRet~$\mathbf{Y}$ 
		}
		
		\tcc{Ego Context Extractor.}
		\Fn{\FECE{$\mathbf{C}^{\sim}_{(\textit{l-1})}$}}{
			$\mathbf{Q}~\leftarrow~\mathbf{C}^{\sim}_{(\textit{l - 1})}$ 
			
			$\mathbf{K}~\leftarrow~\mathbf{V}~\leftarrow~\mathbf{H}^{\sim}_{\Theta} ~\leftarrow~ h_{\Theta}(\mathbf{x}_i, ~\mathbf{x}_j)$
			
			$\mathbf{C}^{\sim}_{(\textit{l})} ~\leftarrow~CMHA(\mathbf{Q}, ~\mathbf{K}, ~\mathbf{V})$ 
			
			\KwRet $\mathbf{C}^{\sim}_{(\textit{l})}$
		}
		
		\tcc{Cross Context Synthesizer.}
		\Fn{\FCCS{$\mathbf{M}^{C}_{(\textit{l-1})},~\mathbf{C}^{A}_{(\textit{l})},~\mathbf{C}^{G}_{(\textit{l})}$}}{
			$\mathbf{Q}~\leftarrow~\mathbf{M}^{C}_{(\textit{l-1})}$ 
			
			$\mathbf{K}~\leftarrow~\mathbf{V}~\leftarrow~\mathbf{C}^{A}_{(\textit{l})} \textcircled{\small{U}} \mathbf{C}^{G}_{(\textit{l})}$ 
			
			$\mathbf{M}^{C}_{(\textit{l})} ~\leftarrow~CMHA(\mathbf{Q}, ~\mathbf{K}, ~\mathbf{V})$ 
			
			\KwRet $\mathbf{M}^{C}_{(\textit{l})}$
		}
		
		\Fn{\FMain}{
			$\mathbf{F}^{\sim} ~\leftarrow ~$ Extract appearance, geometry and content features from $\mathbf{T}$. 
			
			\tcc{Initialization.}
			
			$\mathbf{C}^{\sim}_{(\textit{0})} ~\leftarrow ~ \mathbf{M}^{\sim}_{(\textit{0})} \leftarrow ~ \mathbf{F}^{\sim}$ 
			
			\tcc{Generate collaborative embeddings by NCGM.}
			
			\For{l = 1, 2, 3}{
				
				$\mathbf{C}^{A}_{(\textit{l})} ~\leftarrow~ ECE(\mathbf{C}^{A}_{(\textit{l-1})})$ 
				
				$\mathbf{C}^{G}_{(\textit{l})} ~\leftarrow~ ECE(\mathbf{C}^{G}_{(\textit{l-1})})$ 
				
				$\mathbf{C}^{C}_{(\textit{l})} ~\leftarrow~ ECE(\mathbf{C}^{C}_{(\textit{l-1})})$ 
				
				$\mathbf{M}^{A}_{(\textit{l})} ~\leftarrow~ CCS(\mathbf{M}^{A}_{(\textit{l-1})}, ~\mathbf{C}^{G}_{(\textit{l})}, ~\mathbf{C}^{C}_{(\textit{l})})$ 
				
				$\mathbf{M}^{G}_{(\textit{l})} ~\leftarrow~ CCS(\mathbf{M}^{G}_{(\textit{l-1})}, ~\mathbf{C}^{A}_{(\textit{l})}, ~\mathbf{C}^{C}_{(\textit{l})})$ 
				
				$\mathbf{M}^{C}_{(\textit{l})} ~\leftarrow~ CCS(\mathbf{M}^{C}_{(\textit{l-1})}, ~\mathbf{C}^{A}_{(\textit{l})}, ~\mathbf{C}^{G}_{(\textit{l})})$ 
			}
			
			$\mathbf{E} ~\leftarrow~ \mathbf{M}^{A}_{(\textit{3})} \textcircled{\small{+}} \mathbf{M}^{G}_{(\textit{3})} \textcircled{\small{+}} \mathbf{M}^{C}_{(\textit{3})} $ 
			
			\tcc{Construct pairs.}
			
			$\mathbf{U}~\leftarrow~ Pairing(\mathbf{E})$ 
			
			\If{train}{
				\tcc{Monte Carlo sampling. $S$ is the sample size.}
				
				$ \left[ \mathbf{U}^{S}; \mathbf{GT}_{\sim}^{S} \right] ~\leftarrow~ Sampling(\left[\mathbf{U};\mathbf{GT}_{\sim}\right] , ~S)$ 
				
				\tcc{Separately compute cell/col/row loss.}
				
				$\mathcal{L}_{\sim} ~\leftarrow~ Loss( \mathbf{U}^{S}, ~\mathbf{GT}_{\sim}^{S})$ 
				
				Backward.
			}
			\Else{
				\tcc{Separately predict cell/col/row relationships.}
				
				$\mathbf{F}^{pred}_{\sim}~\leftarrow~ Classify_{\sim}(\mathbf{U})$ 
			}
			\KwRet 
		}
		\caption{NCGM pseudo code.}
		\label{alg}
	\end{algorithm}

	\subsection{Forward Process}

	For clarity, the detailed forward process of NCGM is shown in Alg.~\ref{alg}. Note, the symbol with superscript ``$\sim$'' denotes it is derived from ``appearance'', ``geometry'' or ``content'' modality. And the symbol with subscript ``$\sim$'' represents it belongs to one of ``cell'', "column" or ``row'' relationships. The sample size $S$ of Monte Carlo sampling is set to 10 in the training phase.

	\subsection{Ablation Study of Loss}

	\begin{table}[htp]
		\centering
		\setlength{\tabcolsep}{4.6mm}
		\scalebox{1} {
			\begin{tabular}{cc|ccc}
				\toprule[1.5pt]
				\multicolumn{2}{c|}{Loss Function} & \multicolumn{3}{c}{Setup-B} \\
				\hline
				$\mathcal{L}_{class}$ & $\mathcal{L}_{con}$ & P & R & F1  \\
				\hline
				\cmark & \xmark   & \textbf{98.9} & 98.6 & 98.7     \\
				\xmark & \cmark  & 94.4 & 92.1  & 93.2    \\
				\cmark & \cmark  & {98.8} &	\textbf{99.3} &	\textbf{99.0}  \\
				\bottomrule[1.5pt]
			\end{tabular}
		}
		\caption{Ablation studies of losses on SciTSR-COMP dataset. $\mathcal{L}_{con}$ and $\mathcal{L}_{class}$ are contrastive loss and binary classification loss respectively.}
		\label{Ablation_Loss}
	\end{table}
	We also perform experiments to evaluate the effect of different loss functions. For the sake of fairness, all models with different loss settings are trained with the same backbone model and training data. As shown in Tab.~\ref{Ablation_Loss}, we observe that the model trained by binary classification loss $\mathcal{L}_{class}$ outperforms the one trained by contrastive loss $\mathcal{L}_{con}$, while the combination of $\mathcal{L}_{class}$ and $\mathcal{L}_{con}$ can achieve better performance than either of the two.
	We attribute this to the extra regularization provided by contrastive loss, that makes the model pay more attention to hard negative pairs. As a consequence, our method can learn more discriminative representations of row, column or cell relationships.
	
	\section{Post-processing}
\begin{figure*}[htp]
		\centering
		\includegraphics[width=0.7\linewidth]{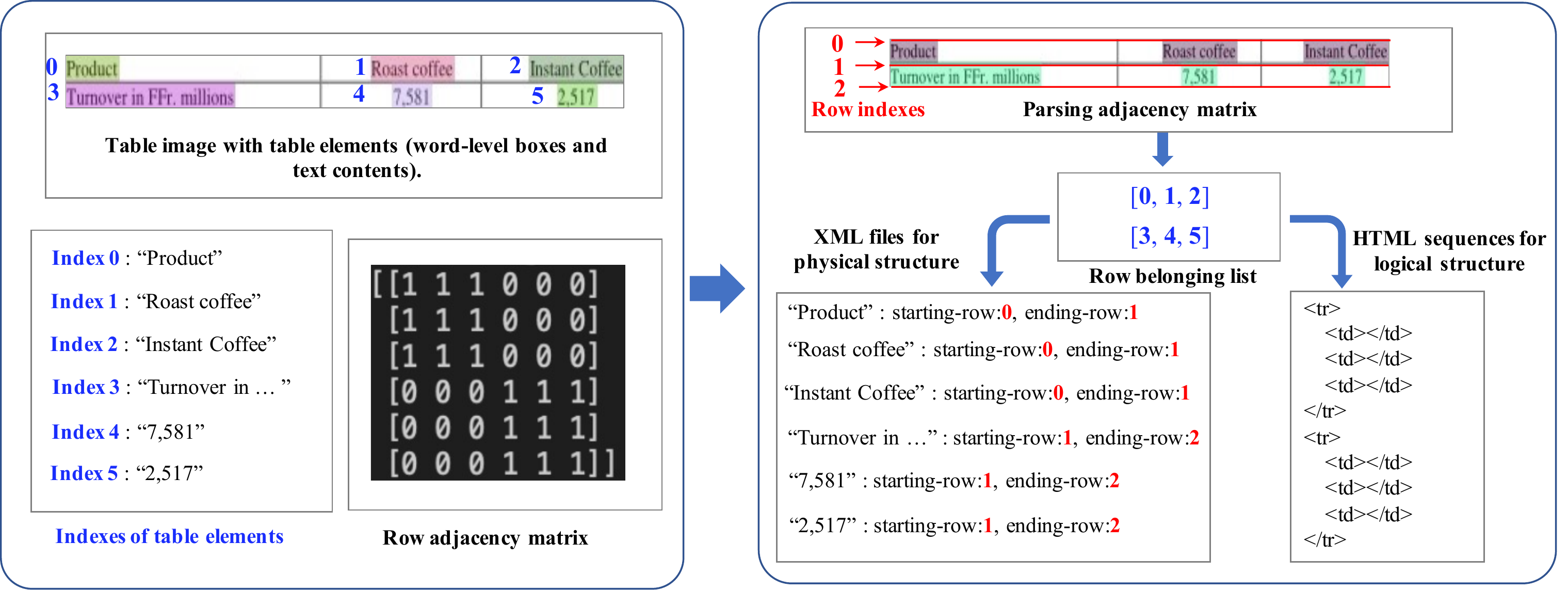}
		\caption{Post-processing of our proposed NCGM.}
		\label{fig:postprocess}
	\end{figure*}

	For a fair comparison with other methods, we perform post-processing on the results of our method. As opposed to pre-processing, post-processing aims to convert the adjacency matrix containing relationships to spanning information either in ``XML'' format for evaluating physical structure recognition or ``HTML'' format for evaluating logical structure recognition respectively, which is shown in Fig.~\ref{fig:postprocess}.
	
	\paragraph{Post-process for physical structure recognition.}
	We also take the row relationship for example. First of all, all boxes are sorted by their $y$ coordinates of top left points to generate their indexes (represented in blue). For each box $v_{i}$, the row belonging list is generated according to row adjacency matrix. Afterwards, the spanning information in ``XML'' format can be obtained. Here, we define the table box row index according to the boundaries of boxes, as illustrated by the red numbers in Fig.~\ref{fig:postprocess}. In detail, boxes belonging to the same row belonging list are assigned with the same starting-row and ending-row indexes. Similarly, we can also obtain the spanning results from column adjacency matrix. Finally, an XML file is created with the extracted spanning information along with bounding box coordinates and contents.
	
	\paragraph{Post-process for logical structure recognition.}
	As for the datasets (\textit{i.e.}, TableBank~\cite{li2020tablebank} and PubTabNet~\cite{end2end1}) in which GTs are in the form of HTML sequences, the evaluation protocol put more emphasis on correctly recognizing the logical structure of tables. We can also convert the adjacency matrix of relationship to HTML tag sequences according to the belonging list. 
	
	\section{Datasets}
	\subsection{Datasets for Experiments}
	We perform large-scale experiments on various benchmark datasets as summarized in Tab.~\ref{Datasets}. Among, 
	ICDAR-2013~\cite{gobel2013icdar}, ICDAR-2019~\cite{gao2019icdar}, UNLV~\cite{shahab2010open}, {WTW~\cite{long2021parsing}},  SciTSR~\cite{chi2019complicated} and SciTSR-COMP~\cite{chi2019complicated} are employed for physical structure recognition, while TableBank~\cite{li2020tablebank} and PubTabNet~\cite{end2end1} are adopted for evaluating logical structure recognition performance. 
	
	In particular, it should be noted that there exists no training set in ICDAR-2013~\cite{gobel2013icdar} and UNLV~\cite{shahab2010open} datasets, so we extend the two datasets to the partial versions~(\textit{i.e.}, ICDAR-2013-P and UNLV-P). Concretely, we randomly split each dataset into five folds, of which four folds for training and the left one for testing. The random splits are performed ten rounds for computing averaged performance, which is similar to TabStruct-Net~ \cite{raja2020table}. 
	
	For more clarity, we also count the number of text segment bounding boxes and tables in every table image for different datasets in Tab.~\ref{Datasets} (``-'' means no training set provided).
	
	\begin{table}[htb]
		\centering
		\setlength{\tabcolsep}{0.6mm}
		\small
		\scalebox{0.9} {
			\begin{tabular}{l|cc|cc|cccc}
				\toprule[1.5pt]
				\multirow{3}{*}{Dataset}  &\multicolumn{2}{c|}{Train} & \multicolumn{2}{c|}{Test} & \multirow{3}{*}{Image} &\multirow{3}{*}{Content} & \multirow{3}{*}{\tabincell{c}{C-Box}} & \multirow{3}{*}{\tabincell{c}{T-Box}} \\
				\cline{2-5}
				& \tabincell{c}{Table\\ (Amt)}   &  \tabincell{c}{Box\\ (Avg)}   & \tabincell{c}{Table\\ (Amt)} & \tabincell{c}{Box\\ (Avg)} &&& \\
				\hline
				IC13 & - & -&158 & 93& \cmark& \cmark& \xmark&  \cmark \\
				IC13-P & 124& 92& 34 &96&  \cmark& \cmark& \xmark& \cmark  \\
				IC19 & 600 &314& 150 &359& \cmark& \xmark&\cmark  &  \xmark   \\
				UNLV & - & -&558 &77& \cmark& \xmark&\cmark    & \xmark  \\
				UNLV-P & 446 &84& 112 &43& \cmark& \xmark&\cmark    & \xmark  \\
				WTW & 10970&101 &3611&96&\cmark &\xmark&\cmark&\xmark \\
				Sci. &12000 &47& 3000 &48& \cmark&\cmark& \xmark&\cmark   \\
				Sci.-C & 12000 &47& 716 &74& \cmark & \cmark & \xmark& \cmark \\
				Sci.-C-A & 24000 &47& 1432 &74& \cmark & \cmark & \xmark& \cmark \\
				TableBank & 145K &50& 1000 &49& \cmark & \xmark&  \xmark& \xmark\\
				PubTabNet & 339K&72 & 114K &74& \cmark & \cmark & \xmark& \cmark \\
				\bottomrule[1.5pt]
			\end{tabular}
		}
		\caption{Statistics of the datasets our experiments performed on. ``Amt'' and ``Avg'' denote ``Amount'' and ``Average'' separately. ``-P'' means partial dataset and ``-A'' represents augmented dataset by distortion. ``IC13'', ``IC19'', ``Sci.'' and ``Sci.-C'' are short for ``ICDAR-2013'', ``ICDAR-2019'', ``SciTSR'' and ``SciTSR-COMP'' individually. ``C-Box'' and ``T-Box'' stand for ``cell bounding boxes'' and ``text segment bounding boxes'' respectively.}
		\label{Datasets}
	\end{table}
	
	\subsection{Processing on Inconsistent Annotation Levels}
	\paragraph{Pre-process for bounding boxes.}
	One major challenge of performing comparisons on different datasets lies in the inconsistency of annotation levels on the bounding boxes.
	As shown in Tab.~\ref{Datasets}, ICDAR-2019~\cite{gao2019icdar}, UNLV~\cite{shahab2010open} and WTW~\cite{long2021parsing} datasets have ground truth~(GT) bounding boxes of cell, while ICDAR-2013~\cite{gobel2013icdar} and SciTSR~\cite{chi2019complicated} datasets take text segment bounding boxes as GT annotations. In our method, we regard text segment bounding boxes as table elements. Therefore, we do some processing to eliminate the inconsistency in annotation levels.
	
	In detail, we convert the cell bounding boxes to the text segment ones according to OCR results in the training stage. For the text-segment-level datasets~(\textit{i.e.}, ICDAR-2013~\cite{gobel2013icdar} and SciTSR~\cite{chi2019complicated}), we consider the original boxes and text contents as model input directly, which are extracted by parsing GT files.
	To unify the input format, for the cell-level datasets~(\textit{i.e.}, ICDAR-2019~\cite{gao2019icdar}, UNLV~\cite{shahab2010open} and WTW~\cite{long2021parsing}), the text-segment-level boxes with contents are generated by the OCR results of Tesseract~\cite{smith2007overview}. Note that an original cell-level box may contain more than one text-segment-level boxes, which have the common row and column spanning information (\textit{i.e.}, starting-row, starting-column, ending-row and ending-column indexes) of the corresponding cell-level box.
	During the testing time, however, we still keep the original cell-level or text-segment-level boxes as GTs instead of the pre-processed ones in Setup-B, which ensures consistency while comparing our method against previously published ones. Especially, we take the result boxes of detection in FLAG-Net~\cite{liu2021show} and the OCR results of Tesseract~\cite{smith2007overview} as inputs for fair comparison in Setup-A.
	
	\vspace{-2mm}
	\paragraph{Pre-process for relationships.}
	In order to provide the uniform GT of adjacency relationships ($\mathbf{GT}_{\sim}$ in Alg.~\ref{alg}) for the model's training phase, we convert the spanning information of table's rows and columns in various formats into the adjacency matrices of cell, row and column, which represent three adjacency relationships for the table elements.
	Take the row adjacency matrix for example, if the $i$-th and $j$-th boxes belong to the same row relationship, the value located at $(i,j)$ in adjacency matrix is assigned to 1, otherwise to 0. In this way, we can construct the row adjacency matrix to represent the relationship of row. The adjacency matrices of cell and column are also generated in the similar way.

	\subsection{Synthesizing Method}
\begin{figure}[htb]
		\centering
		\subfloat[Original Image]{\includegraphics[width=0.32\columnwidth, height=1.8cm]{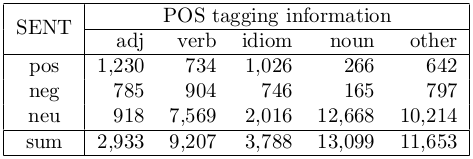}} 
		\subfloat[Distortion 1]{\includegraphics[width=0.32\columnwidth, height=1.8cm]{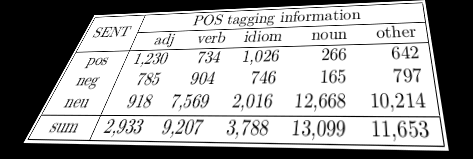}}
		\subfloat[Distortion 2]{\includegraphics[width=0.32\columnwidth, height=1.8cm]{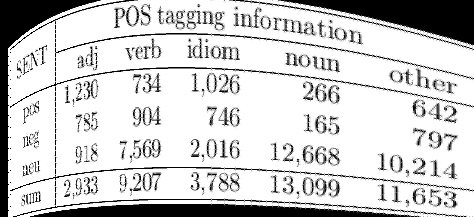}}
		\caption{Images from SciTSR-COMP dataset applied by distortion algorithms.}
		\label{fig:aug} 
	\end{figure}

	To further investigate the capacity of TSR methods under more challenging scenes, we augment existing datasets with the following two kinds of image distortion algorithms to simulate distractors brought by capture device, which are visualized in Fig.~\ref{fig:aug}.
	\paragraph{Distortion 1.}
	The first disortion is based on perspective transformation algorithm, which projects the table image to a new view plane according to the mapping matrix, as is shown in Fig.~\ref{fig:aug}(b).
	\vspace{-2mm}
	\paragraph{Distortion 2.}
	For the second kind of distortion, we employ a algorithm based on the quadratic Bézier curve~\cite{joy2000quadratic} to augment the datasets, which can be defined as:
	\begin{align*}
		\small
		B_{2}(t) = (1-t)^{2}P_{0} + 2t(1-t)P_{1} + t^{2}P_{2}, t\in[0,1], 
	\end{align*} 
	where $P_{0}$,  $P_{1}$ and $P_{2}$ denote three control points of the Bézier curve. 
	
	\vspace{-2mm}
	\begin{figure}[htb]
		\centering
		\includegraphics[width=0.7\columnwidth]{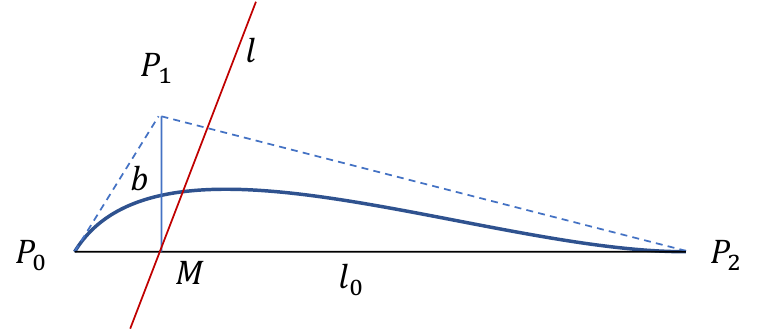}
		\caption{Determination of control points in Bézier curve.} 
		\label{fig:distortion}
	\end{figure}
	\vspace{-2mm}
	Concretely, for each row of the image, we generate a quadratic Bézier curve applied on it to implement pixel-level distortion. There are three main steps to determine the control points of quadratic Bézier curve. As shown in Fig.~\ref{fig:distortion}, we first randomly initialize the axis line $l$ (the red line) and the offset $b$. Next, each row of the image is regarded as $l_{0}$,  and its starting point is deemed as the control point $P_{0}$ while ending point as $P_{2}$. Besides, the control point $P_{1}$ is located at a position offset from $M$~(the intersection point between $l_{0}$ and $l$) by $b$. Through this way, the quadratic Bézier curves are determined by the control points, which are applied on each row of image pixels to perform distortion.
	It is worth mentioning that the blank pixels generated in the distortion process are interpolated by neighbouring pixels.

	\section{Computational Complexity} 
	To further compare the computational complexity of existing various methods of table structure recognition, we summarize the model sizes and the inference operations of different models in Tab.~\ref{param}.
	Since LGMPA~\cite{qiao2021lgpma} and Cycle-CenterNet~\cite{long2021parsing} recover table structure based on heuristic rules after detecting cells, which is infeasible to perform the comparison between them and our method, we do not report them in Tab.~\ref{param}.
	In particular, note that TabStruct-Net~\cite{raja2020table} and FLAG-Net~\cite{liu2021show} are only tested for structure recognition, so we do not count the parameters and operations of cell detection for a fair comparison.
	
	Although the parameters and FLOPs of NCGM are larger than FLAG-Net~\cite{liu2021show}, the performance of our method increases average F1-score by a large margin especially under challenging scenarios~(\textit{e.g.}, WTW and SciTSR-COMP-A). 
	The reasons for increasing computational complexity is probably because of the individual operations on multiple modalities in our method. 
	Compared with TabStruct-Net~\cite{raja2020table}, NCGM can achieve better performance with less parameters and similar computational budgets.
	Moreover, the model size and FLOPs of GraphTSR~\cite{chi2019complicated} are the smallest among the compared methods, but it only utilizes the box coordinates as input to recognize table structure, which cannot achieve comparable performance than other methods. We consider to optimize the computational complexity and size of model without performance degradation in the future work.

	\begin{table}[htb]
		\setlength{\tabcolsep}{5.8mm}
		\centering
		\scalebox{1} {
			\begin{tabular}{l|cc}
				\toprule[1.5pt]
				\multirow{2}{*}{Method} & \multicolumn{2}{c}{Setup-B} \\
				\cline{2-3} 
				& \#Param& FLOPs \\
				\hline
				\tabincell{l}{GraphTSR~\cite{chi2019complicated}} & 7.0e-4 & 1.8e-4 \\
				\tabincell{l}{DGCNN~\cite{rethinkingGraphs}} & 0.8 & 4.1 \\
				\tabincell{l}{TabStruct-Net~\cite{raja2020table}} & 4.7 & 11.9 \\
				FLAG-Net~\cite{liu2021show}&1.9 & 3.3 \\
				\hline
				NCGM & 3.1 & 12.7 \\ 
				\bottomrule[1.5pt]
			\end{tabular}
		}
		\caption{Computational complexity comparison of different methods. \#Param denotes the number of parameters (M), while FLOPs are the numbers of FLoating point OPerations (G). The number of input table's text segment bounding boxes is 42.}
		\label{param}
	\end{table}

	\section{Jensen-Shannon Divergence}
\begin{figure*}[htb!]
		\centering
		\subfloat[Block1]{\includegraphics[width=0.32\linewidth,height=3.6cm]{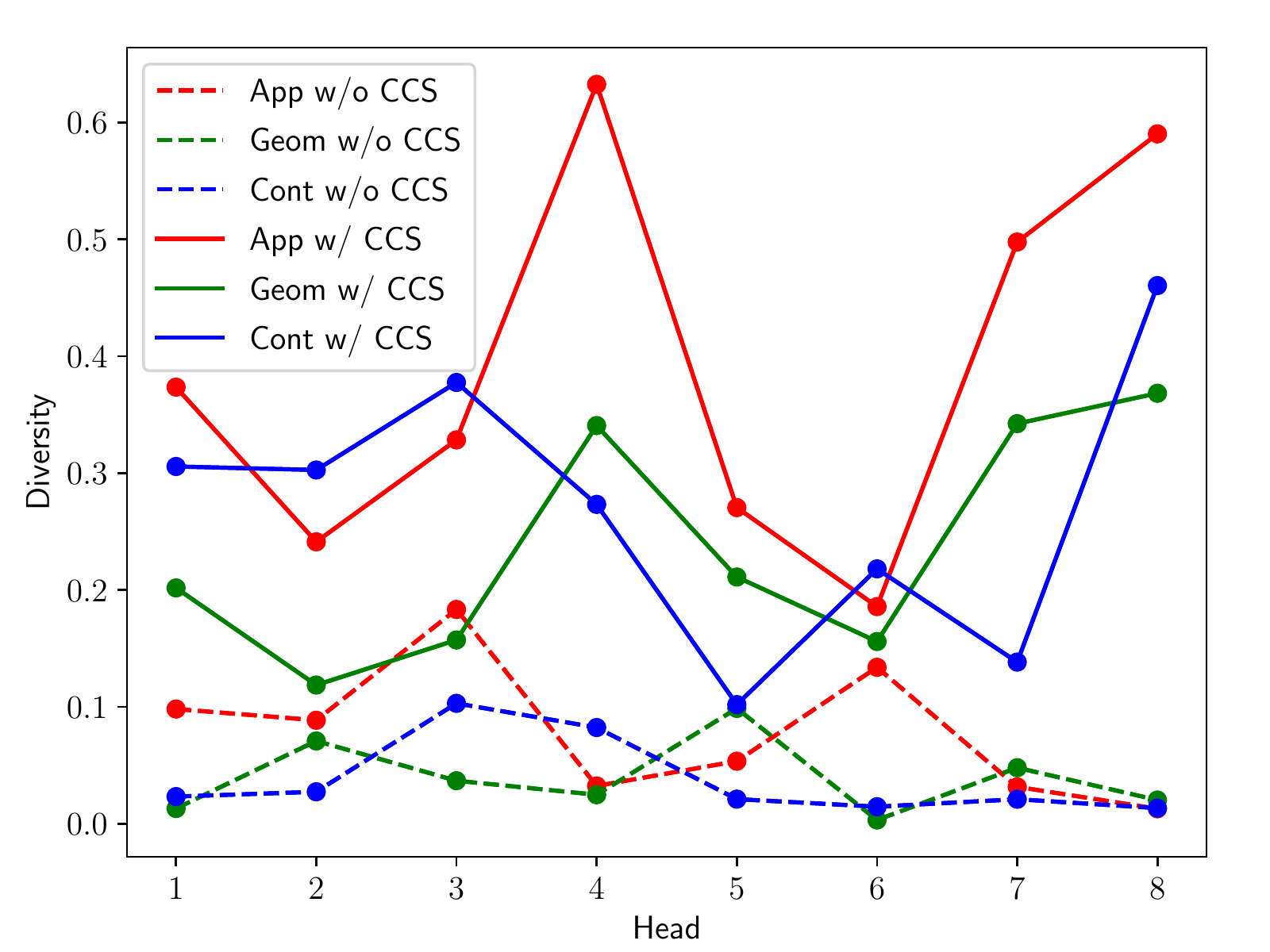}} \hspace{2mm}
		\subfloat[Block2]{\includegraphics[width=0.32\linewidth,height=3.6cm]{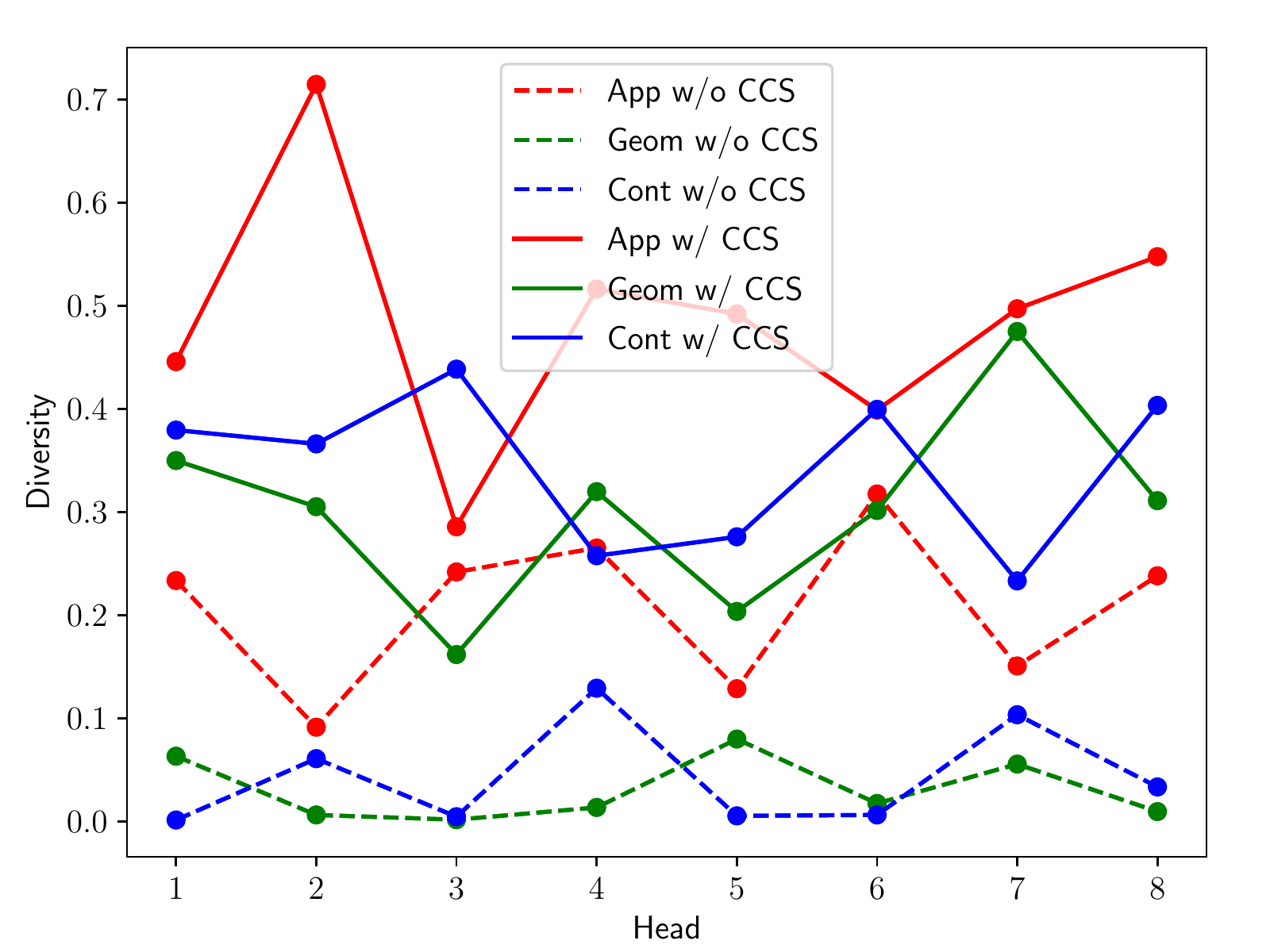}}\hspace{2mm}
		\subfloat[Block3]{\includegraphics[width=0.32\linewidth,height=3.6cm]{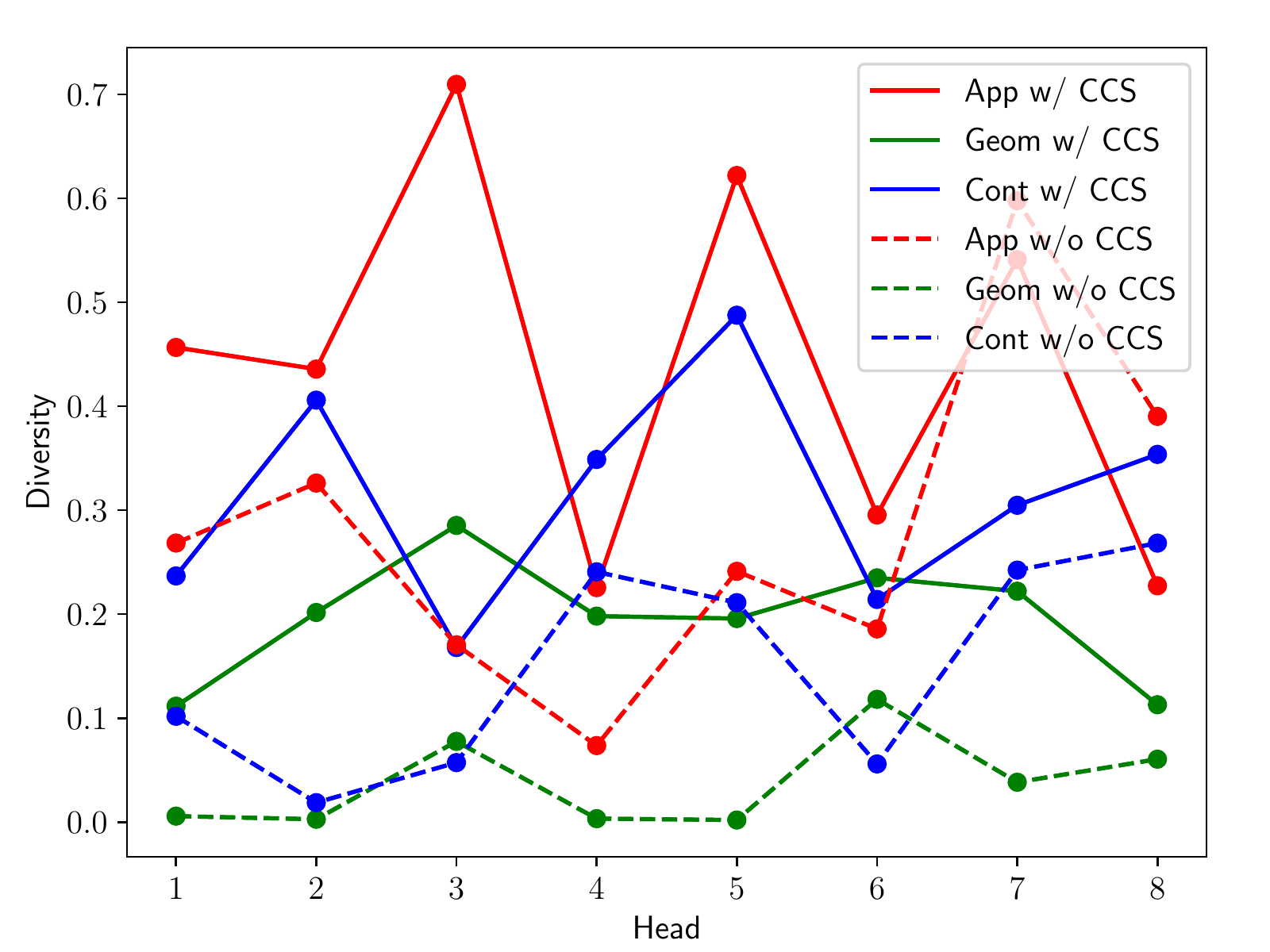}}
		\vspace{-0.2cm}
		\caption{Diversities of attention maps for different modalities with or without CCS in different blocks.} 
		\label{fig:CCS} 
		\vspace{-0.4cm}
	\end{figure*}

	We in this work introduce the Jensen-Shannon Divergence~\cite{correia2019adaptively} to measure the average diversity of attention maps in CCS, which is defined as:
	\begin{align*}
		\small
		JSD = H(\frac{1}{n}\sum_{i=1}^{n}\mathbf{P}_{i}) - \frac{1}{n}\sum_{i=1}^{n}H(\mathbf{P}_{i}), 
	\end{align*} 
	where $\mathbf{P}_{i}$ is the vector of attention weights assigned by one head to $i$-th node in the graph, and
	$H$ is the Shannon entropy. The trends of attention diversity variance in different blocks for different modalities with and without CCS are all shown in Fig.~\ref{fig:CCS}.

	\section{Qualitative Results}
	
\begin{figure*}[htb!]
		\centering
		\subfloat[Sample result of NCGM on ICDAR-2013 dataset.]{\includegraphics[width=0.3\linewidth, height=2cm]{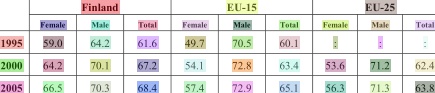}\hspace{4mm}\includegraphics[width=0.3\linewidth, height=2cm]{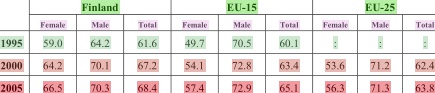}\hspace{4mm}\includegraphics[width=0.3\linewidth, height=2cm]{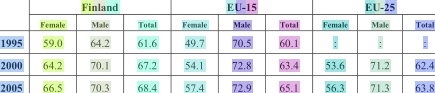}}

		\subfloat[Sample result of NCGM on ICDAR-2019 dataset.]{\includegraphics[width=0.3\linewidth, height=2.5cm]{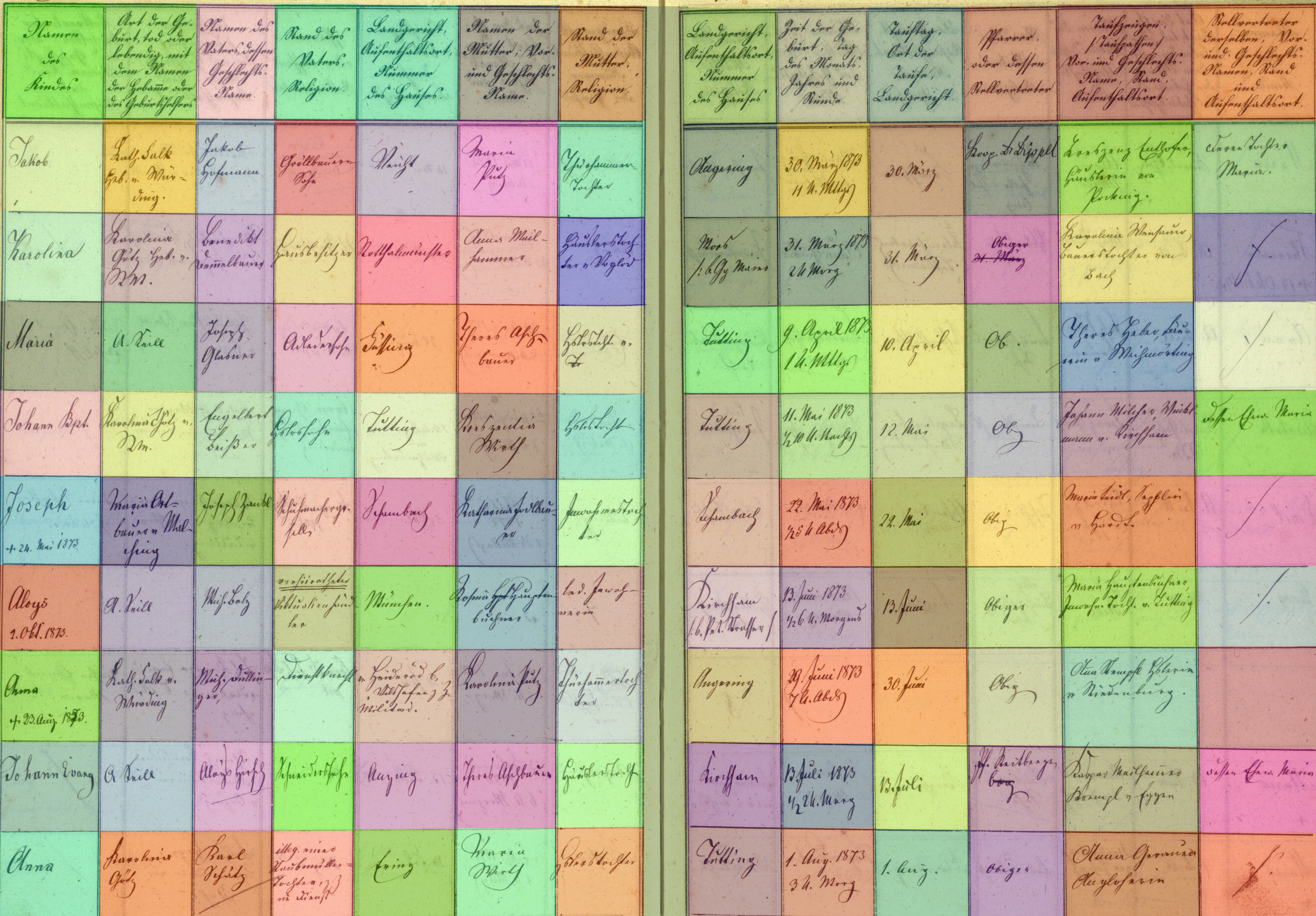}\hspace{4mm}\includegraphics[width=0.3\linewidth, height=2.5cm]{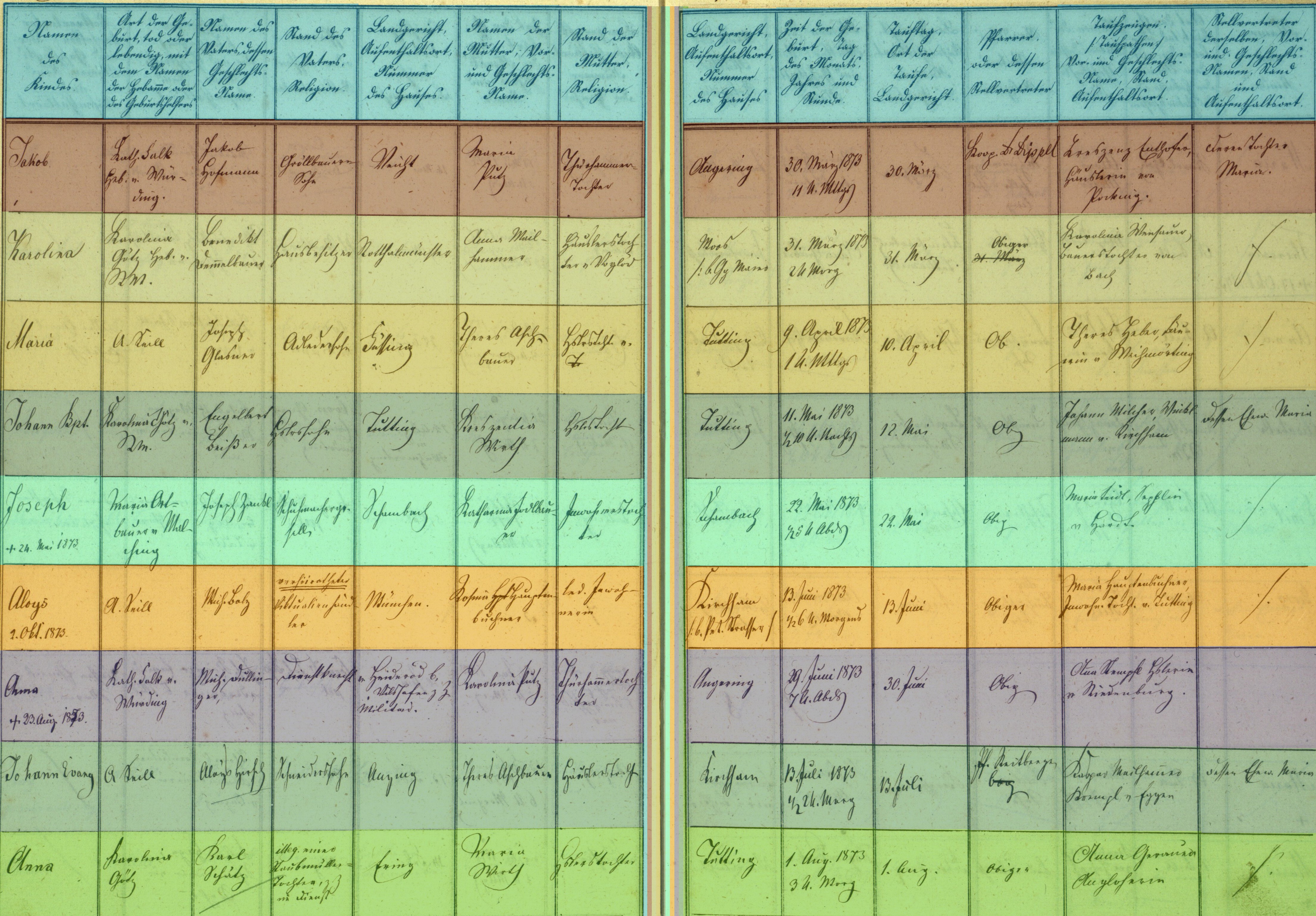}\hspace{4mm}\includegraphics[width=0.3\linewidth, height=2.5cm]{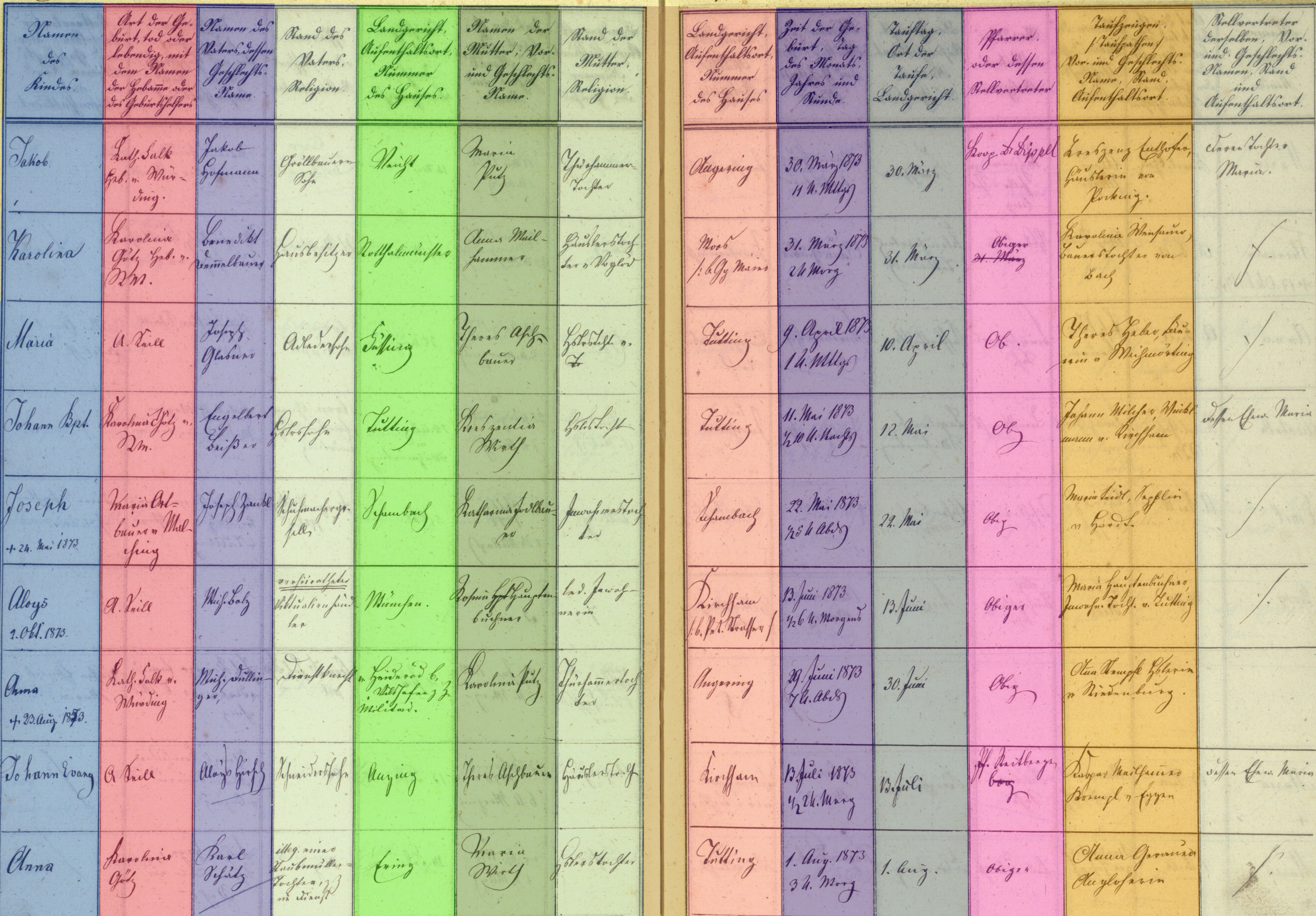}}

		\subfloat[Sample result of NCGM on UNLV dataset.]{\includegraphics[width=0.3\linewidth, height=2.5cm]{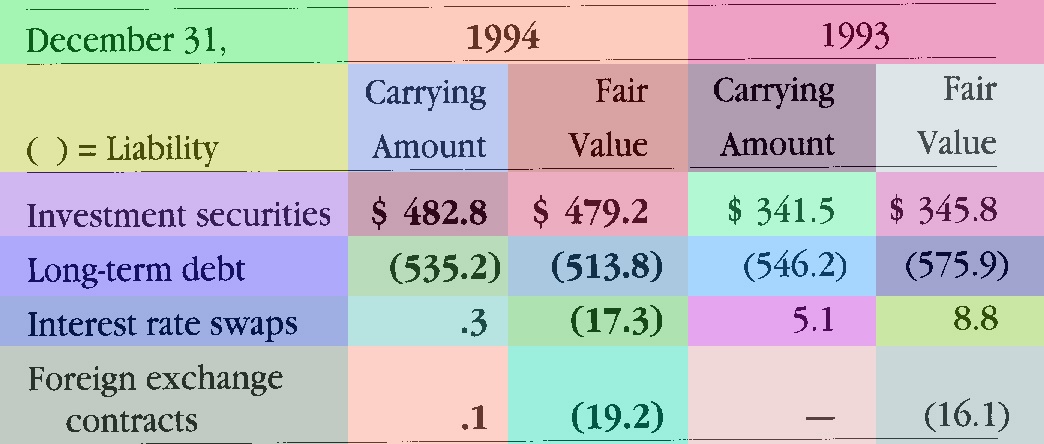}\hspace{4mm}\includegraphics[width=0.3\linewidth, height=2.5cm]{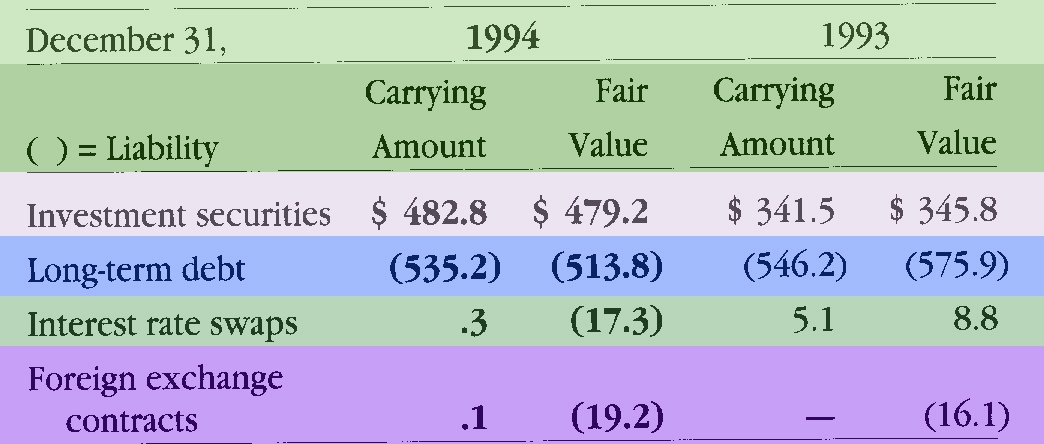}\hspace{4mm}\includegraphics[width=0.3\linewidth, height=2.5cm]{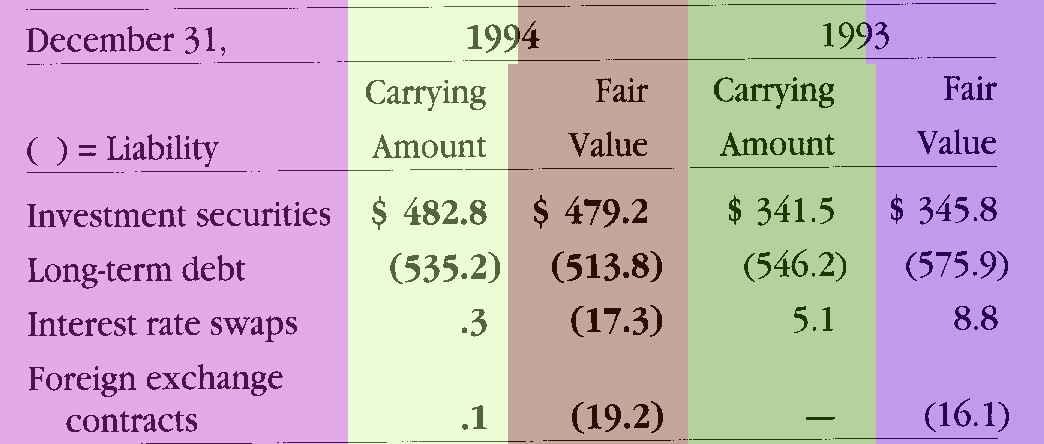}}

		\subfloat[Sample result of NCGM on SciTSR dataset.]{\includegraphics[width=0.3\linewidth, height=2.5cm]{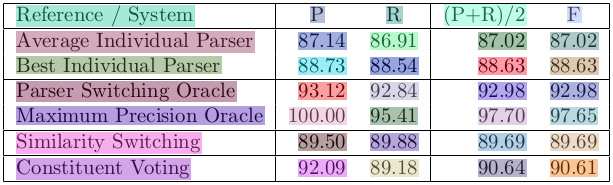}\hspace{4mm}\includegraphics[width=0.3\linewidth, height=2.5cm]{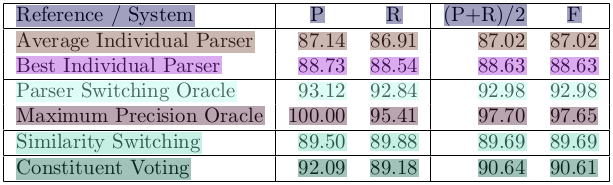}\hspace{4mm}\includegraphics[width=0.3\linewidth, height=2.5cm]{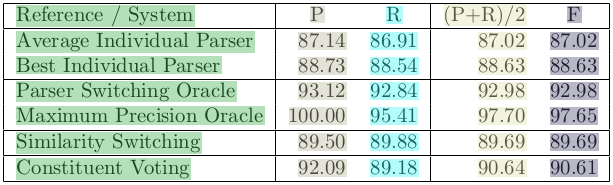}}

		\subfloat[Sample result of NCGM on SciTSR-COMP dataset.]{\includegraphics[width=0.3\linewidth, height=2.5cm]{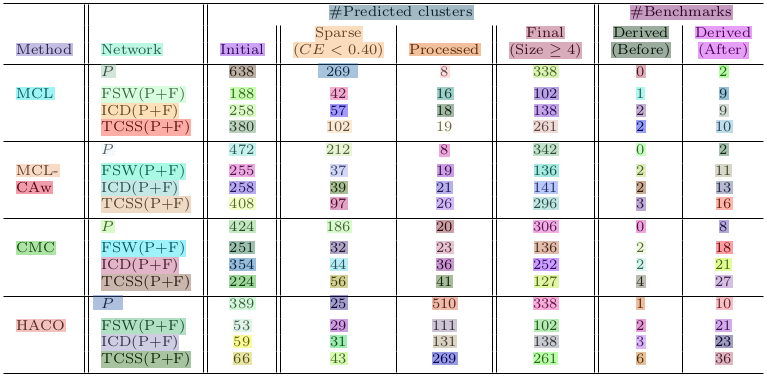}\hspace{4mm}\includegraphics[width=0.3\linewidth, height=2.5cm]{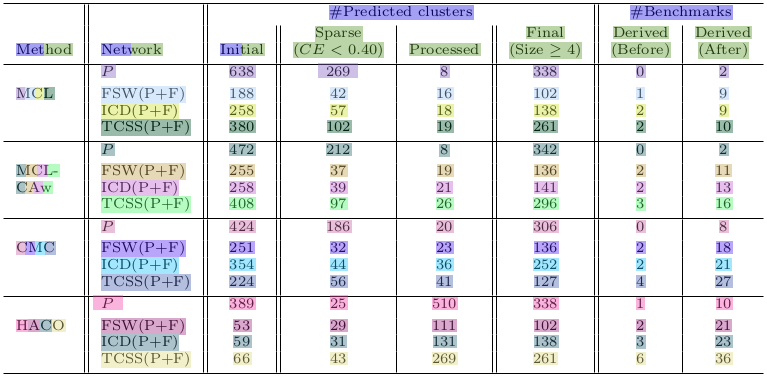}\hspace{4mm}\includegraphics[width=0.3\linewidth, height=2.5cm]{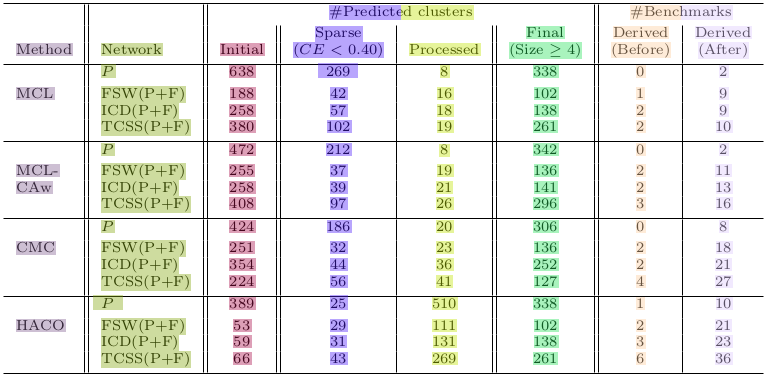}}

		\subfloat[Sample result of NCGM on SciTSR-COMP-A~(Distortion 1) dataset.]{\includegraphics[width=0.3\linewidth, height=3cm]{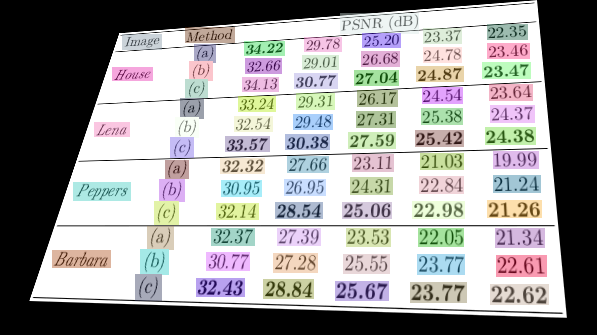}\hspace{4mm}\includegraphics[width=0.3\linewidth, height=3cm]{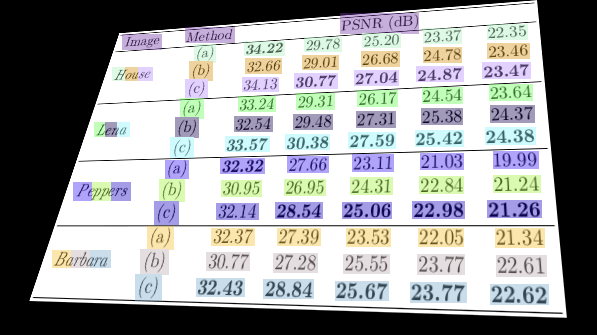}\hspace{4mm}\includegraphics[width=0.3\linewidth, height=3cm]{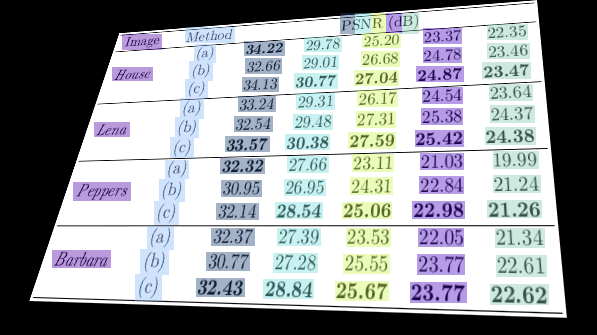}}

		\subfloat[Sample result of NCGM on SciTSR-COMP-A~(Distortion 2) dataset.]{\includegraphics[width=0.3\linewidth, height=3cm]{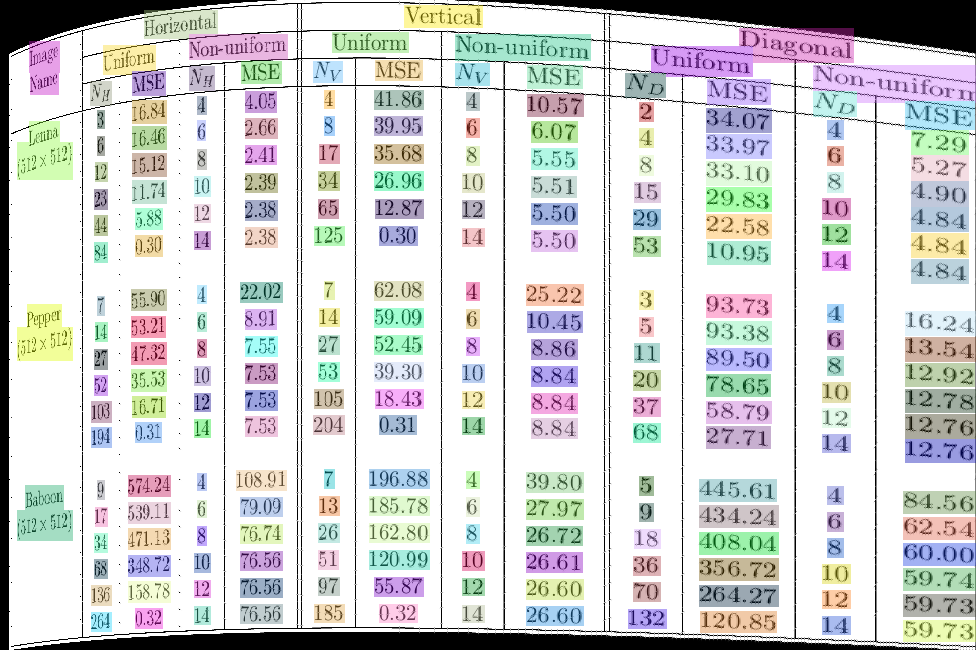}\hspace{4mm}\includegraphics[width=0.3\linewidth, height=3cm]{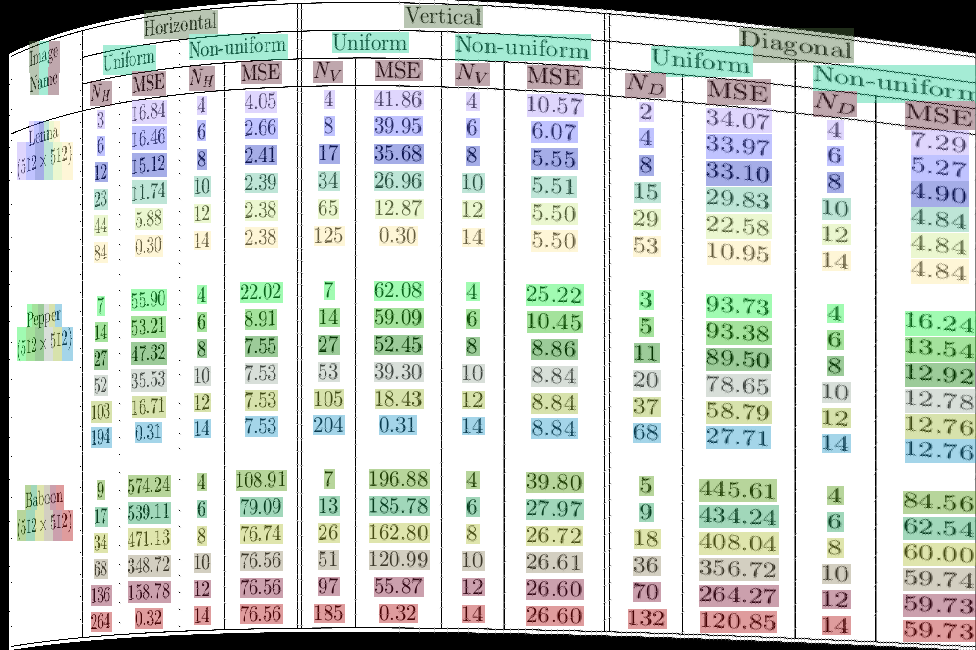}\hspace{4mm}\includegraphics[width=0.3\linewidth, height=3cm]{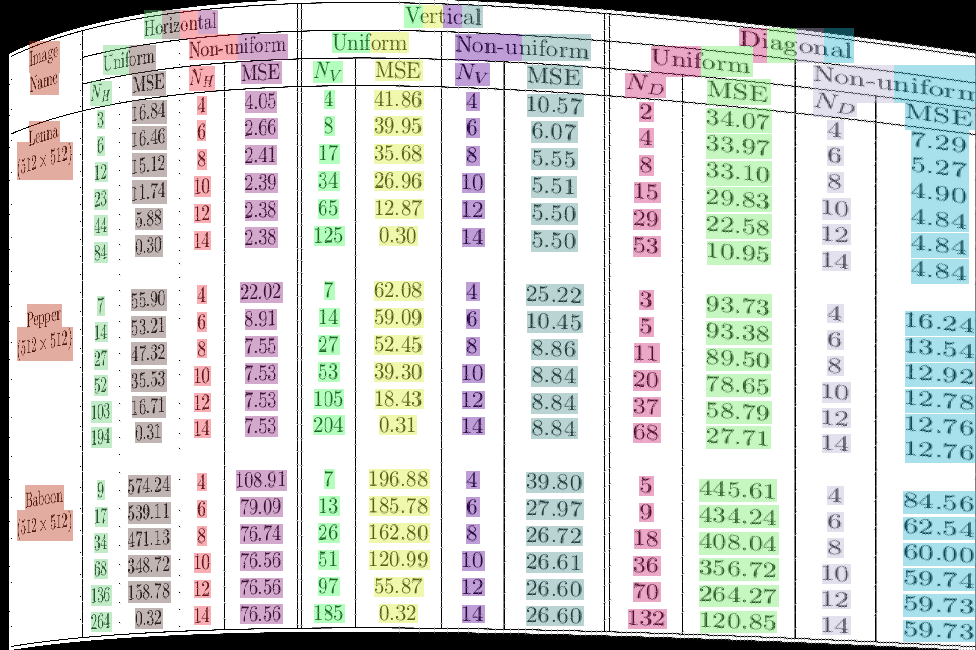}} 
		\caption{Sample TSR output of NCGM on table images of various datasets. The first, second and last column indicate the
			predictions of cells, rows and columns respectively.} 
		\label{fig:sample}  
	\end{figure*}

	Fig.~\ref{fig:sample} demonstrates more qualitative results of structure recognition on benchmark datasets. The figures show the generalization ability of our proposed NCGM which is able to correctly recognize various types of table structures. Especially for more challenging cases, Fig.~\ref{fig:sample}(f)-(g) verify that our method can not only handle regular tables
	but also robustly recognize distorted ones, which is more applicable in realistic scenarios.

\begin{figure*}[htb!]
		\centering
		\subfloat[Cell Relationships]{\includegraphics[width=0.8\linewidth]{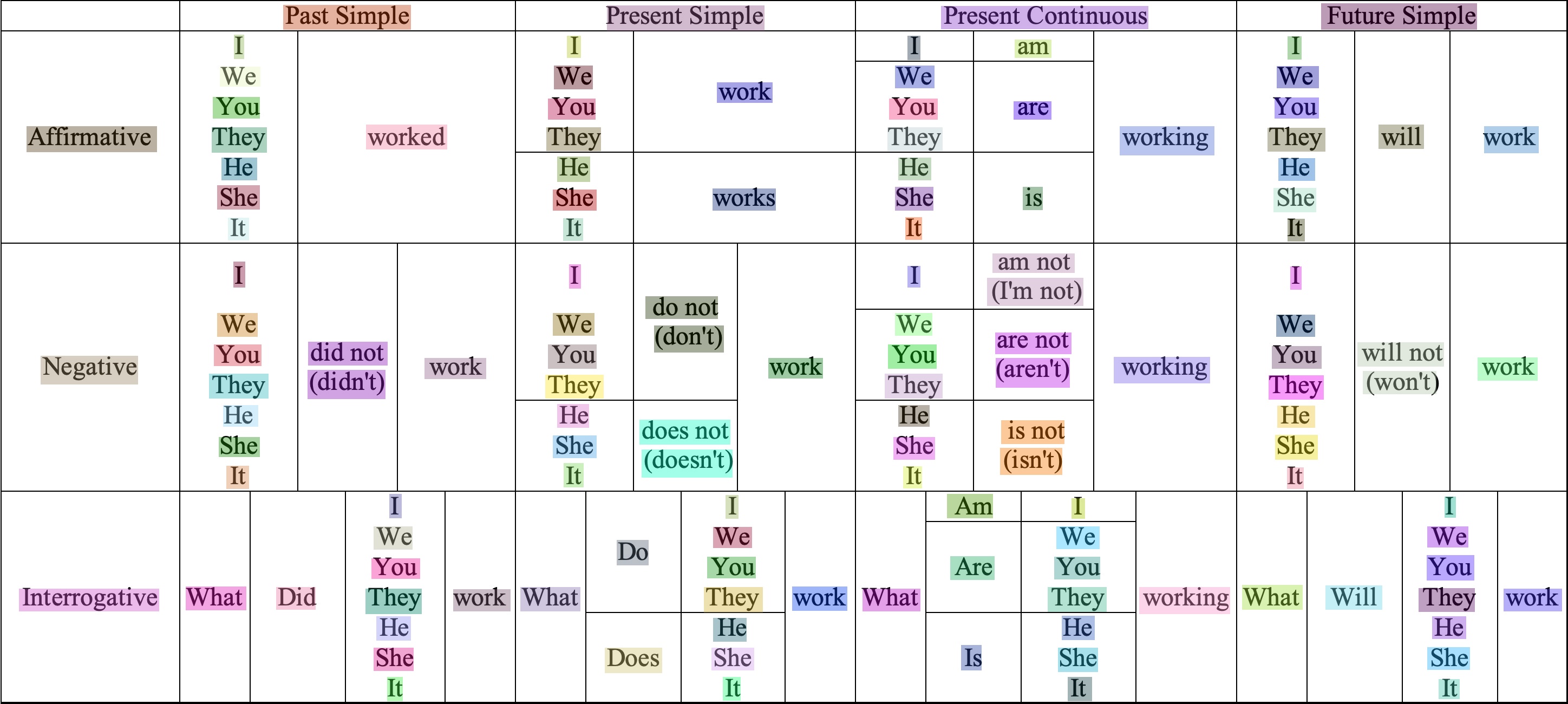}} 
		
		\subfloat[Row Relationships]{\includegraphics[width=0.8\linewidth]{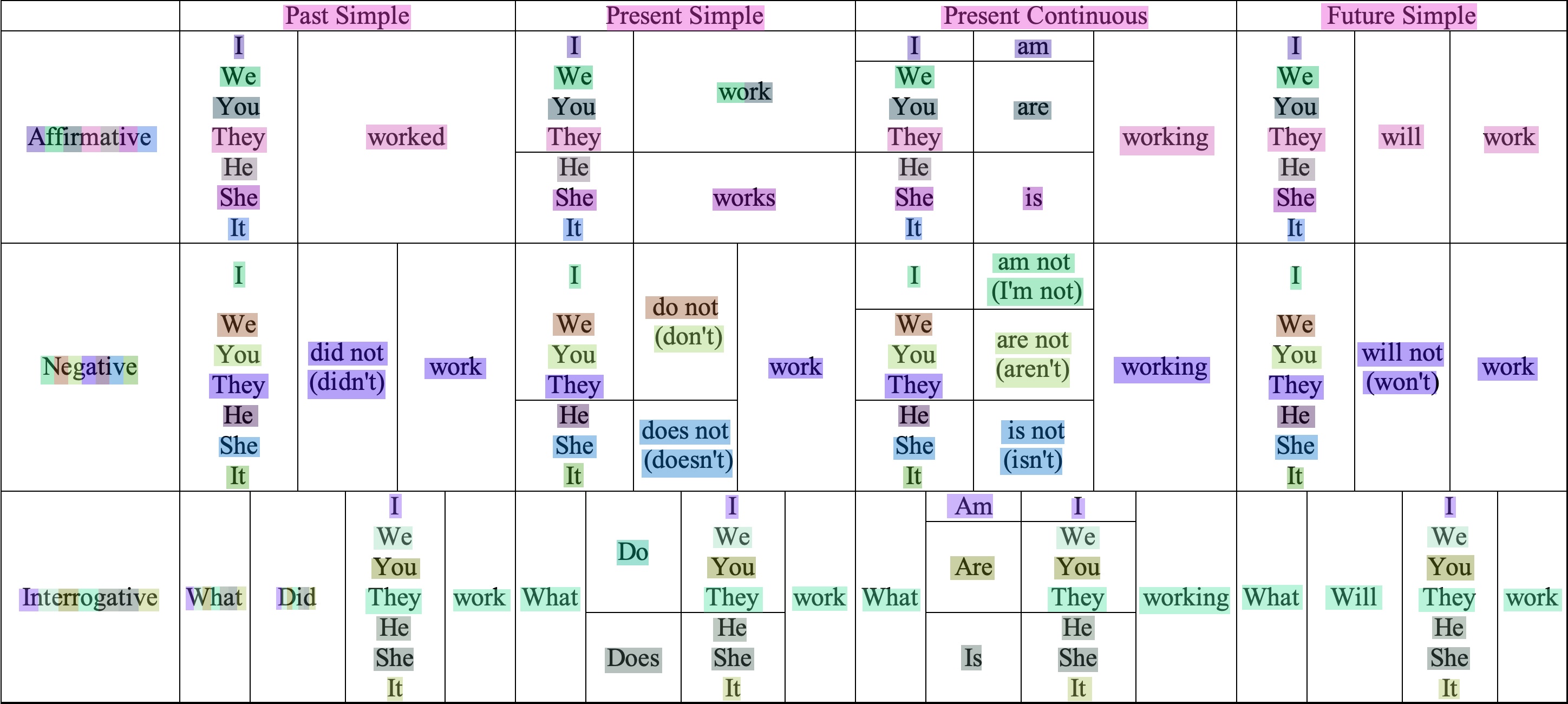}}
		
		\subfloat[Column Relationships]{\includegraphics[width=0.8\linewidth]{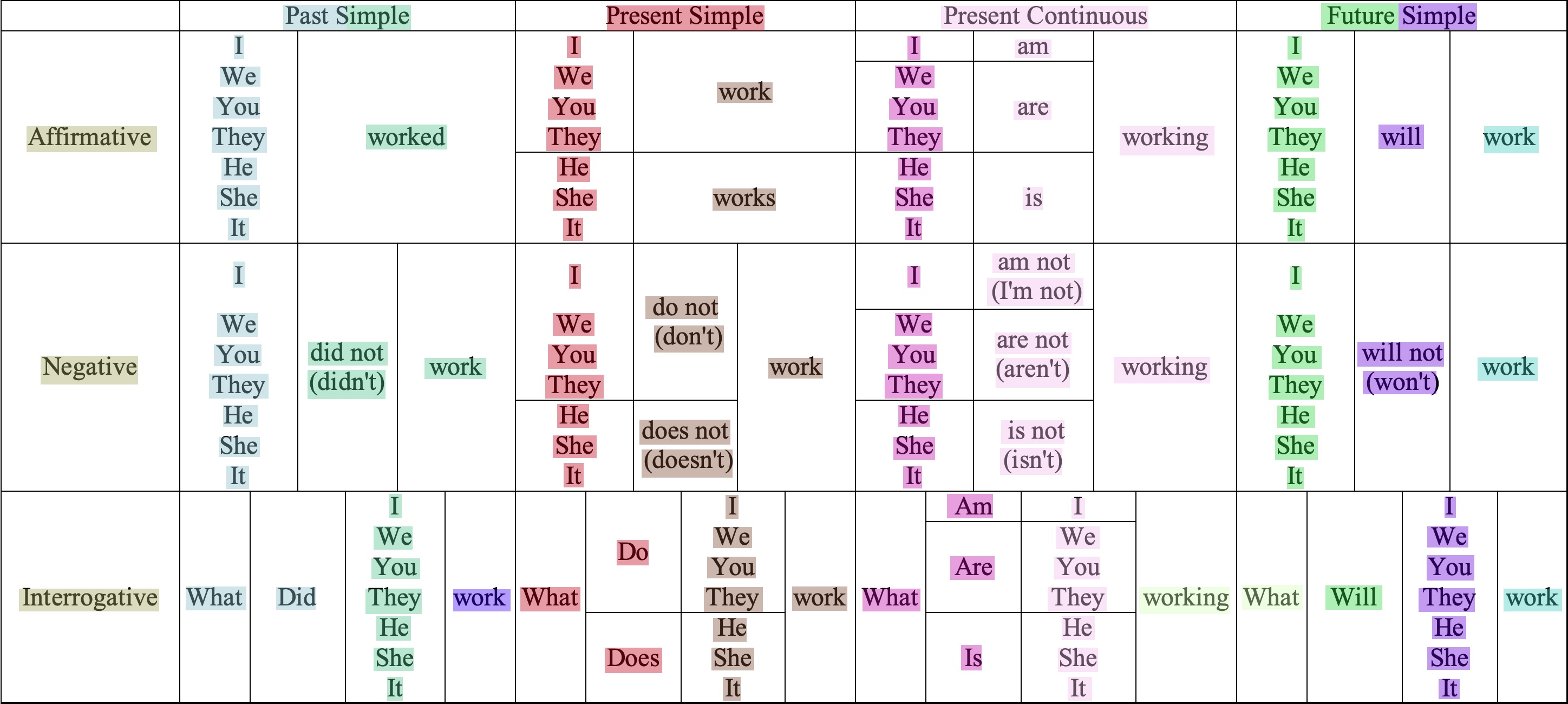}}
		\caption{Failure cases of NCGM on table with more complex structure.}
		\label{fig:sample_n}
	\end{figure*}

	We also show the failure cases of our method in Fig.~\ref{fig:sample_n}. As one can see, the table that impairs the performance of our algorithm is the nested table, which contains severe misalignment of row and column. To put it in another way, it is ambiguous to judge whether certain boxes belong to the same row or column. The ambiguity also incurs inadaptability of existing evaluation protocols in either logical or physical format requiring the rigid alignment of box boundary in row or column relationships. In the future work, we will investigate this problem and attempt to attack it by introducing more robust representation of the nested table structure, such as tree structure.

	\section{Broader Impact}
	Table elements have natural graph structure. Learning collaborative patterns from graph data of multiple modalities offers many potential applications and opportunities as graph data in multiple modalities naturally co-occur and have implicit relationships. Our model can be applied in many specific verticals ranging from financial area to medical area including large-scale heterogeneous table data, such as financial documents, medical examination reports and \textit{etc}. And we focus on the impact our model might have on them. A model that is capable of dealing with large-scale multi-modality data is extremely significant for table information registration and data analysis. With the development of smart phones, a large amount of table images are captured by mobile cameras in realistic application. Different from regular table images obtained by scanner or parsing PDF metadata, those captured by mobile device contain more distractors (\textit{e.g., }distortion).  Table structure recognition~(TSR) algorithm plays as the front-end role that converts input table image to machine readable data, which is vital to the whole document processing system. However, most of existing TSR methods are merely designed for regular tables and cannot generate satisfactory results from table cases with more challenging distractors. Thanks to the more effective capture of inter-intra modality interaction, our model tailored for Hetero-TSR can yield more precise results, especially under more challenging scenarios, which is demonstrated by extensive experiments. In other words, our model can not only greatly save labor costs and improve document processing efficiency, but show more extensibility in application scenarios. Besides, we provide a successful attempt in the direction of investigating the collaborative patterns with and between modalities. We encourage researchers to build graph embedding models based on NCGM for other graph-based tasks we can expect to be particularly beneficial.    
\end{appendices}

{\small
\bibliographystyle{ieee_fullname}
\bibliography{table_bib}
}
\end{document}